\pdfoutput=1
\documentclass[11pt,onecolumn]{cleantechnicalreport}

\usepackage[authoryear,sort&compress,round]{natbib}
\usepackage{hybridfrontmatter}
\usepackage{nicefrac}
\usepackage{flafter}
\usepackage{placeins}
\usepackage{float}
\usepackage{makecell,multirow}
\usepackage{longtable}
\usepackage[noabbrev]{cleveref}
\newcommand{\ResourceMedianCells}{90}
\newcommand{\ResourceMeanAboveMedianCells}{89}
\newcommand{\ResourceFullRows}{111}

\newcommand{\PaperTrajectoryCount}{52,168}

\title{When Agents Fail: Model or Harness—Who’s to Blame?}
\renewcommand{\ReportTitleLogo}{%
  \includegraphics[width=19mm,trim=40bp 108bp 61bp 155bp,clip]{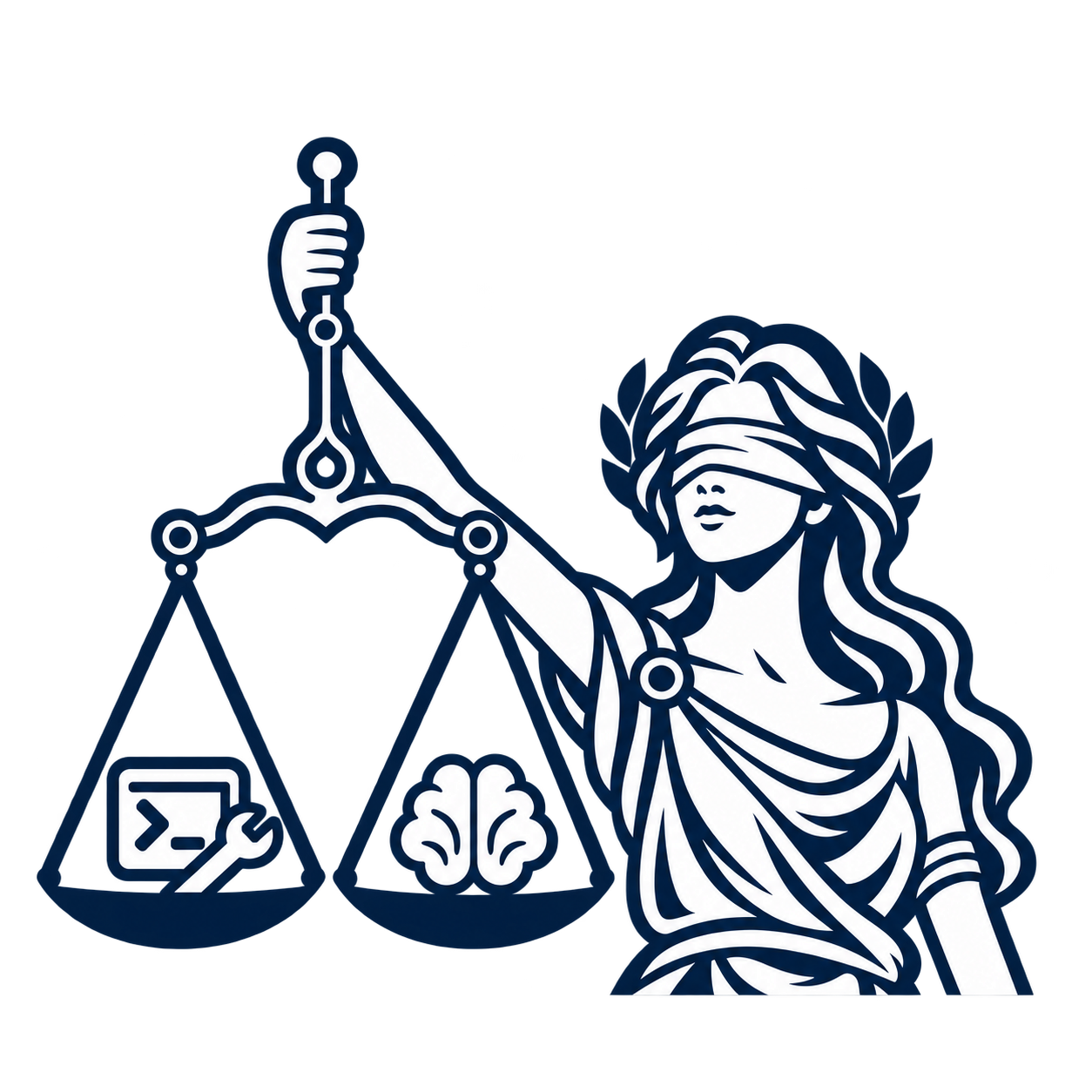}%
}
\renewcommand{\ReportTitleText}{%
  \begin{tabular}[b]{@{}l@{}}
    {\normalfont\bfseries\fontsize{21}{24}\selectfont\color{ReportTheme}When Agents Fail:}\\[1.2mm]
    {\normalfont\fontsize{16}{20}\selectfont Model or Harness—Who’s to Blame?}
  \end{tabular}%
}

\renewcommand{\ReportAuthorFont}{\normalfont\fontsize{10.5}{14}\selectfont}
\author{Shao Zhang}
\author{Hangfan Zhang}
\author{Yuanyuan Shi}
\author{Hao Li}
\author{Min-Le Su}
\author{Kangcong Li}
\author{Jingping~Wang}
\author{Chunjiang Mu}
\author{Chen Zhang}
\author{Yang Chen}
\author{Bo Zhang}
\author{Shuyue~Hu
}
\affil{Shanghai Artificial Intelligence Laboratory\\
\texttt{\{zhangshao,hushuyue\}@pjlab.org.cn}}

\DeclareUnicodeCharacter{2212}{-}

\newcommand{\OverviewTeaser}{%
  \noindent\begin{minipage}{\linewidth}
    \centering
    \captionsetup{type=figure,font=footnotesize,skip=3pt,hypcap=false}
    \begin{subfigure}[b]{.44\linewidth}
      \centering
      \includegraphics[width=\linewidth,trim=0 8.64bp 0 0,clip]{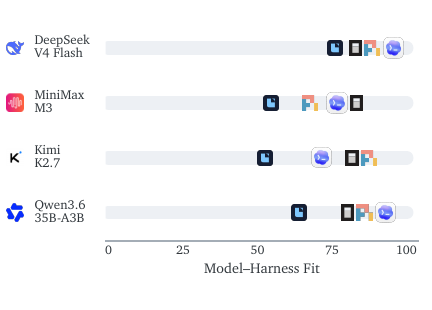}
      \caption{}
    \end{subfigure}%
    \hfill
    \begin{subfigure}[b]{.24\linewidth}
      \centering
      \includegraphics[width=\linewidth,trim=0 8.64bp 0 0,clip]{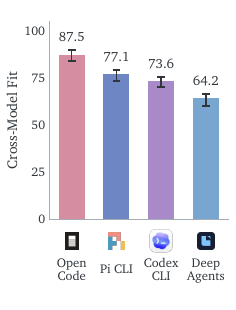}
      \caption{}
    \end{subfigure}%
    \hfill
    \begin{subfigure}[b]{.30\linewidth}
      \centering
      \includegraphics[width=\linewidth,trim=0 8.64bp 0 0,clip]{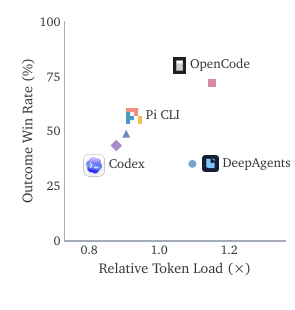}
      \caption{}
    \end{subfigure}%
    \caption{Overview of Model-Harness Fitness.
    (a) Model-specific fit captures task achievement through valid completion;
    DeepSeek has the highest average Fit across harnesses.
    (b) Cross-model fit summarizes overall harness suitability across models;
    OpenCode ranks first.
    (c) Task performance versus token consumption: greater token use does not
    necessarily correspond to better outcomes.
    Panel (a) uses four models on four shared benchmarks
    (SWE-bench-45, Terminal-Bench 2, VitaBench, and LOCA-Bench);
    (b,c) use three models on six benchmarks.}
    \label{fig:teaser-overview}
  \end{minipage}\par\vspace{3mm}%
}
\pretocmd{\abscontent}{\OverviewTeaser}{}{%
  \PackageError{overview-teaser}{Cannot insert teaser before abstract}{}%
}

\begin{abstract}
LLM agents combine models with execution harnesses, so agentic benchmark performance reflects their joint behavior. We argue that harnesses should therefore be evaluated through their interactions with underlying models. We develop a framework combining five complementary measures: (i) Valid Completion Rate (VCR), how reliably agents finish normally and deliver evaluable results, irrespective of task correctness; (ii) Model--Harness Fit, how well a harness supports task achievement through valid completion for a given model; (iii) Cross-Model Fit, a harness's overall suitability across evaluated models; (iv) Interaction Residuals, pairing-specific execution strengths and weaknesses hidden by component-level summaries; and (v) Token Load, the token demands of agent execution, including failed attempts. We evaluate four general-purpose harnesses (\textit{OpenCode}, \textit{Codex CLI}, \textit{Pi CLI}, and \textit{DeepAgents}) with four core models, plus two fixed native model--harness systems, on six benchmarks covering context management, tool use, long-context planning, multimodal interaction, and artifact handoff. Including native references and stress conditions, \textbf{111 evaluation settings} yield \textbf{\PaperTrajectoryCount{} trajectories}. Harness rankings vary across models and benchmarks; task performance and valid completion can favor different pairings, and similar task scores can accompany distinct execution profiles. \textbf{OpenCode ranks first in cross-model performance, while DeepSeek V4 Flash Preview maintains consistently high Mean VCR across harnesses on the shared four-benchmark panel}. These findings support evaluating model--harness interactions, training transferable harness-use capabilities in models, and harness designs adapted to the model and task.
\end{abstract}

\begin{document}
\maketitle

% \notes{current harness}

% \notes{why we need harness Eval}
% efficency/model capability/ harness with model problem

% \notes{what we do - for model bench to harness}
% adapter/eval metrics

% \notes{experiments}
\section{Introduction}

\begingroup
\setlength{\intextsep}{6pt}
\begin{figure}[!t]
\centering
\setlength{\abovecaptionskip}{4pt}
\includegraphics[width=\textwidth]{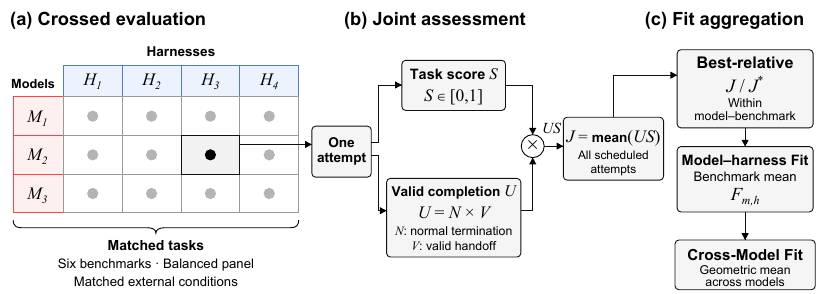}
\caption{Overview of the model--harness evaluation framework.}
\label{fig:evaluation-overview}
\end{figure}
\endgroup

Classical views describe intelligent agents as goal-oriented systems that continuously perceive their environment, maintain internal state, select actions, and improve behavior through feedback \citep{wooldridge1995intelligent,wang2024survey}. An LLM agent implements this paradigm by integrating a large language model (LLM) into an executable system for perception, reasoning, and action in external environments \citep{yao2023react}. At the system level, an LLM agent combines its model with an execution layer, the \emph{agent harness}, that connects the model to the environment \citep{guo2026harness,kwa2025timehorizon,yao2026harnessbench}. The harness provides mechanisms for observation processing, context management, tool and action interfaces, workflow control, state management, execution, evaluation, and error handling. By controlling how information is presented to the model and how its outputs are executed, the harness shapes agent behavior and task performance \citep{zhang2025dpt}.

Under this system view, end-to-end benchmark performance reflects the joint behavior of the model--harness system rather than model capability alone \citep{yang2024sweagent,yao2026harnessbench}. Yet most benchmarks remain task-centric, evaluating fixed model--harness configurations by task success or aggregate reward while treating the harness as part of the setup. These evaluations summarize system-level success; execution diagnostics and crossed
comparisons provide complementary views of performance sources. Failures may reflect insufficient model capability, ineffective harness use, or harness limitations in observation processing, action interfaces, state management, or recovery \citep{ma2024agentboard,zhu2026unified,yao2026harnessbench}. An empirical issue study reports failures occurring only when particular model behaviors interact with harness parsing, control flow, or context handling \citep{chen2026agentreactive}. Conversely, high performance may partly reflect the harness's task-specific workflows or additional computation \citep{lou2026autoharness,lee2026metaharness}. Controlled studies show that changing harnesses can alter performance and rankings with models and tasks fixed \citep{zhu2026unified,yao2026harnessbench}. Understanding agent failure therefore requires examining the model, the harness, and their particular pairing through crossed comparisons and execution evidence. This assessment distinguishes failure to solve the task from abnormal termination or invalid evaluator handoff: a run may complete validly without solving the task, or earn task credit despite abnormal termination.

The missing object of evaluation lies at the model--harness boundary. A model's \emph{harness-use capability} depends on how effectively it can interact with a given harness: the same interface, action space, or feedback mechanism may be easy for one model to use but challenging for another, while the same model may behave differently under different harness designs \citep{yang2024sweagent,yao2026harnessbench}. Accordingly, harness difficulty should be understood as a model-dependent interaction cost introduced by the harness's observation format, action interface, state management, control flow, and recovery mechanisms. When this cost exceeds the model's harness-use capability, the harness may prevent the model from realizing its underlying capabilities; when the harness provides extensive task-specific support, it may improve performance by contributing part of the task-solving process. Existing task-level evaluations conflate these effects into a single outcome, making it difficult to distinguish model capability from harness effects \citep{messick1995validity,zhu2026unified,yao2026harnessbench}. Therefore, evaluating agents requires examining both task achievement and valid completion, together with how these outcomes vary across model--harness pairings.

We develop metrics to analyze model--harness interaction through
\emph{task performance} and \emph{operational validity}. Valid Completion Rate
(VCR) measures how often attempts terminate normally with valid evaluator
handoff, regardless of task correctness. Model--harness Fit and Cross-Model Fit
summarize task performance from valid completions for individual models and
across models. We analyze their diagnostic value and robustness, failure
patterns, and resource use. We implement a common \emph{agent adapter} for
this study, preserving native execution loops under matched tasks and
controlled execution conditions. We cross four general-purpose
harnesses---DeepAgents, OpenCode, Pi CLI, and Codex CLI---with multiple models
on six benchmarks: SWE-bench-45
\citep{jimenez2024swebench,openai2024swebenchverified}, Terminal-Bench 2
\citep{merrill2026terminalbench}, VitaBench \citep{he2026vitabench},
LOCA-Bench \citep{zeng2026locabench}, ClawMark \citep{meng2026clawmark}, and
AgentSkillOS \citep{li2026agentskillos}. Fixed native systems add complete-system
references. \Cref{fig:evaluation-overview} summarizes the evaluation framework.

Harness rankings vary across models and benchmarks in both task performance
and operational validity, which can favor different harnesses for one model.
Their joint distribution distinguishes execution profiles among configurations with
similar success rates: clean runs may leave tasks unsolved, while non-clean runs may
succeed or earn continuous-score credit. Pair-specific VCR
deviations beyond model and harness averages also change sign across benchmarks.
Resource use varies across models and tasks: the lightest harness changes across
configurations, and lower token use can accompany lower outcomes. Low use can reflect
premature termination; high use can include repeated failed actions.
Cross-Model Fit identifies a stable aggregate leader on the evaluated panel,
but model-specific preferences and intermediate ranks depend on its models
and benchmarks.

\Needspace{3\baselineskip}
These findings motivate joint assessment of task performance and operational
validity: their joint distribution identifies task progress achieved through
valid completion, while resource use describes its cost. Harness use should
be trained and evaluated as a transferable capability across interfaces,
including action construction, feedback use, recovery, and stopping. Harness
design should adapt observation, recovery, and resource-control policies to
the model and task.

\section{Related Work}
\label{gen_inst}

\paragraph{Harness design and optimization.}
LLM agents couple a model to an execution layer that constructs observations,
exposes actions, maintains state, and closes the feedback loop. ReAct made this
loop explicit, while SWE-agent and OpenHands showed how interfaces and runtime
infrastructure shape agent behavior
\citep{yao2023react,yang2024sweagent,wang2025openhands}. A parallel line treats
the surrounding system as an optimization target: GPTSwarm optimizes agent
graphs, and ADAS and AFlow search over code-defined agent systems and workflows
\citep{zhuge2024gptswarm,hu2025adas,zhang2025aflow}. More recent methods
optimize constraints, context, full harness implementations, or editable agent
code from execution feedback
\citep{lou2026autoharness,lee2026metaharness,xu2026lifeharness,
zhang2026selfharness,zhang2026dgm,lin2026ahe}. These methods show that the
surrounding program can substantially change fixed-model performance. These methods focus on agent-system optimization. Evaluating interaction difficulty,
task-solving structure, and pairing-specific compatibility is a complementary objective.

\paragraph{Harness evaluation.}
Most agent benchmarks report task outcomes for a complete system
\citep{liu2024agentbench,mialon2024gaia,yao2025taubench,fu2025catarena}; AgentBoard adds
progress and trajectory diagnostics but remains task-centered
\citep{ma2024agentboard}. Recent work addresses execution-layer confounding by
crossing native harnesses and models or using shared adapters with standardized
traces \citep{yao2026harnessbench,zheng2026clawswebench,wang2026a2e}.
Harness-Bench and harness-evolution analyses further show that harness effects
are model-dependent
\citep{zhang2026selfharness}. Other studies
recommend reporting harness--model pairs, resource use, and failure profiles,
and matching feedback budgets and held-out tasks
\citep{vats2026scaffoldeffect,wang2026rethinkingharness}. Trace-level scaling
work further distinguishes useful feedback from raw resource expenditure
\citep{zhang2026efc}. These studies complement final-success measurements with rankings and diagnostics of
complete configurations. Our
work measures task performance and operational validity across model--harness
pairings, evaluates their joint contribution to effective operation, and
quantifies pairing-specific deviations from additive component averages.

\section{Preliminaries}
\label{sec:preliminaries}

Interpreting agent scores requires specifying the model, harness, and task
conditions that produced them \citep{messick1995validity}. We first describe
how a model and harness form an executable agent, then distinguish task
performance from operational validity and explain how their joint assessment
characterizes effective operation. We also establish conditions for comparing
model--harness pairings.

\begingroup
\setlength{\intextsep}{6pt}
\begin{figure}[!t]
\centering
\setlength{\abovecaptionskip}{4pt}
\includegraphics[width=\textwidth,trim={0 140bp 0 100bp},clip]{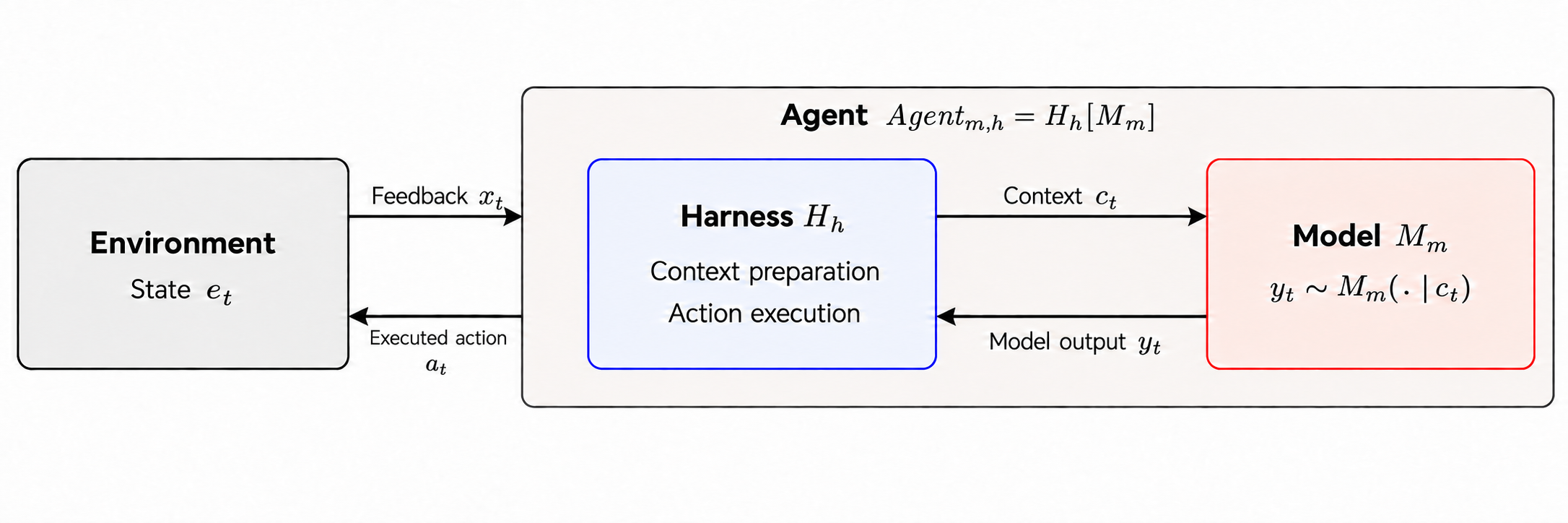}
\caption{Model--harness composition and execution loop. The harness prepares
model context and executes approved actions; model outputs $y_t$ and executed
actions $a_t$ are distinct.}
\label{fig:model-harness-formulation}
\end{figure}
\endgroup

\subsection{Model--Harness Configurations}

An LLM agent combines a language model with an execution harness to interact
with an external environment, which holds task state such as repository
files or application records. The model proposes text or tool use; the
\emph{agent harness} prepares model context and manages action execution
\citep{guo2026harness,macedo2026whatharness}. Model proposals are distinct
from executed actions, whose results and errors feed back into subsequent
steps \citep{yao2023react,yang2024sweagent}. For a model $M_m$ and harness
$H_h$, indexed by $m$ and $h$, the executable agent is
$\mathsf{Agent}_{m,h}=H_h[M_m]$. The environment and task success criteria
remain outside the harness.
\Cref{fig:model-harness-formulation} summarizes this loop;
\Cref{app:harness-formalization} defines its interfaces.

\subsection{Formal Runtime Interfaces}
\label{app:harness-formalization}

At interaction step $t$, the environment has state $e_t$, exposes a raw
signal $x_t\sim\Omega(\,\cdot\mid e_t\,)$, accepts an executable action $a_t$,
and evolves as $e_{t+1}\sim T(\,\cdot\mid e_t,a_t\,)$, where $\Omega$ and $T$
are the observation and transition kernels. This partially observed
interaction model \citep{kaelbling1998planning} specifies the environment
interface. The model $M_m$ is a conditional generator whose output proposal
$y_t$ may contain text, structured tool-use proposals, or control suggestions.

We define the harness as a stateful runtime with six interfaces:
\[
H_h=\langle
\operatorname{Observe}_h,\operatorname{Context}_h,
\operatorname{Control}_h,\operatorname{Act}_h,
\operatorname{Update}_h,\operatorname{Govern}_h
\rangle.
\]
$\operatorname{Observe}_h$ converts raw feedback into model-usable
observations; $\operatorname{Context}_h$ selects and formats instructions,
history, tool descriptions, memory, and artifacts;
$\operatorname{Control}_h$ coordinates generation, retries, recovery,
delegation, requests for input, and stopping; $\operatorname{Act}_h$ maps model proposals to tool
or environment operations; $\operatorname{Update}_h$ persists runtime memory
and artifacts; and $\operatorname{Govern}_h$ enforces permissions,
budgets, validation, verification, and handoff. This decomposition treats the
harness as a runtime substrate rather than a prompt wrapper
\citep{guo2026harness,macedo2026whatharness}.

The harness maintains \emph{runtime memory} $\mu_t$: selected interaction
information retained across steps as summaries, memory entries, artifact
references, and execution bookkeeping. This implementation-level record is
distinct from the external environment state $e_t$. For task goal $g$,
the observation, context, and generation interfaces are
\[
o_t=\operatorname{Observe}_h(x_t,\mu_t),\qquad
c_t=\operatorname{Context}_h(g,o_{\leq t},\mu_t),\qquad
y_t\sim M_m(\,\cdot\mid c_t\,).
\]
Here $o_{\leq t}$ denotes the observation history through step $t$. The harness
interprets and governs $y_t$, executes approved actions, and incorporates
environment feedback into memory updates $\mu_t\to\mu_{t+1}$. A model invocation may
lead to zero or more executed actions, whereas the environment transition
equation describes a single action $a_t$. Repeated and parallel
generations, subagents, persistent memory, and adaptive runtime memory are all
instances of this same composition.

\subsection{Operational Dimensions and Controlled Comparisons}
Building on this execution loop, we distinguish two dimensions of
model--harness operation: \emph{task performance}, how well the final state
satisfies the task's success criterion, and \emph{operational validity},
whether execution terminates normally with an evaluator-consumable state,
regardless of correctness. The same interface can impose different demands
across models and tasks \citep{cronbach1972dependability}. Task success and
intermediate-action correctness describe the task-solving process, and task
credit may survive abnormal termination. We therefore assess the two dimensions
separately for diagnosis and jointly to characterize \emph{effective operation}:
task performance achieved through valid completion of the same attempt.

Crossed comparisons vary the model and harness independently, holding one fixed
while changing the other. Task instances and external conditions, including
tool access and resource limits, are matched to separate pairing effects from
changes in task conditions \citep{fisher1935design}. Each harness retains its native execution loop and bundled design choices, so the unit
of comparison is the complete model--harness system \citep{holland1986statistics}.

\section{Metric Design}
\label{sec:metric-design}

We first measure task performance and operational validity, then compare
harnesses on each dimension and jointly through Fit. Interaction residuals
describe VCR deviations beyond model and harness averages, while resource
metrics quantify execution cost. Cross-benchmark summaries
use a \emph{balanced panel}: the same models and harnesses evaluated in every
combination on all included benchmarks. The panel and repetition count used
in this study are specified in \Cref{sec:experiment-setup}.

\paragraph{Task performance and operational validity.}

Official evaluator outcomes, averaged over $R$ repetitions, measure task performance.
For operational validity on task $i$ and repetition $r$, let $N_{i,r}$ indicate normal termination and $V_{i,r}$
an evaluator-consumable submission or final state. A \emph{clean run} achieves
valid completion, $U_{i,r}=N_{i,r}V_{i,r}=1$, irrespective of correctness. For
each model--harness pair with benchmark task set $\mathcal T$, the
\emph{Valid Completion Rate} (VCR) is
\begin{equation}
\mathrm{VCR}
=\frac{1}{|\mathcal T|\,R}
\sum_{i\in\mathcal T}\sum_{r=1}^{R}U_{i,r}.
\end{equation}
\emph{Mean VCR} averages these rates equally across the included
benchmarks for each model--harness pair. We report VCR and Mean VCR as
percentages. \Cref{app:execution-stage-decomposition} separates abnormal termination from
invalid handoff after normal termination.

\paragraph{Outcome Win Rate and VCR Win Rate.}

To compare each dimension across benchmarks, we score harness $h$ against
$h'\ne h$ within each model--benchmark block using the indicator function
$\mathbf{1}(\cdot)$:
\begin{equation*}
w_{h,h'}=\mathbf{1}(s_h>s_{h'})+0.5\cdot\mathbf{1}(s_h=s_{h'}).
\end{equation*}
Here, $s_h$ is the primary outcome for \emph{Outcome Win Rate} or VCR for
\emph{VCR Win Rate}. Win rates summarize pairwise ordering across benchmarks;
aggregation and reporting conventions are given in
\Cref{app:scoring-aggregation}.

\paragraph{Model--harness Fit and Cross-Model Fit.}
For a joint assessment, Fit compares task scores earned through valid
completion across harnesses for the same model. Let $S_{i,r}^{m,h,b}\in[0,1]$ be the normalized
official task score for model $m$, harness $h$, and benchmark $b$, retaining
partial credit. Valid attempts contribute their task scores; invalid attempts contribute zero. The mean over all scheduled attempts is
\begin{equation}
\label{eq:joint-valid-score}
J_{m,h,b}=\frac{1}{|\mathcal T_b|R}
\sum_{i\in\mathcal T_b}\sum_{r=1}^{R}U_{i,r}^{m,h,b}S_{i,r}^{m,h,b}.
\end{equation}
For each model and benchmark, let $J^\star_{m,b}=\max_{h'}J_{m,h',b}$ be the
best observed harness score in the candidate panel \citep{dolan2002benchmarking}.
Model--harness Fit, reported as $100F_{m,h}$, averages $J_{m,h,b}/J^\star_{m,b}$
equally over benchmarks $\mathcal B$; Cross-Model Fit takes the
geometric mean across evaluated models $\mathcal M$:
\begin{equation}
\label{eq:cross-model-fit}
F_{m,h}=\frac{1}{|\mathcal B|}\sum_{b\in\mathcal B}
\frac{J_{m,h,b}}{J^\star_{m,b}},
\qquad
\mathrm{Cross\text{-}Model\ Fit}(h)
=100\left(\prod_{m\in\mathcal M}F_{m,h}\right)^{1/|\mathcal M|}.
\end{equation}
Both indices range from 0 to 100. For example, a model--harness Fit of 80
means reaching 80\% of the best observed
score from valid completions for that model, averaged equally across benchmarks.
Cross-Model Fit ranks harnesses across the evaluated models; its geometric
mean penalizes low model-specific Fit \citep{fan2023welfare}.

\paragraph{Joint-score interpretation and aggregation.}
For binary benchmarks, $J$ is the fraction of scheduled attempts that both
succeed and achieve valid completion; for continuous benchmarks, it is the
mean task-score contribution from valid completions. When VCR is positive,
$J=\mathrm{VCR}\times\mathbb E[S\mid U=1]$, using VCR as a fraction.
The identity $\mathbb
E[S]=\mathbb E[US]+\mathbb E[(1-U)S]$ decomposes the official score into contributions
from valid completions and other attempts.

\paragraph{Interaction residuals.}

Let $v_{m,h}$ denote Mean VCR (\%) on the balanced panel. We quantify
panel-dependent deviations from additive model and harness averages in
percentage points as
\begin{equation}
\label{eq:pairing-affinity}
\Delta_{m,h}=v_{m,h}-v_{m,\cdot}-v_{\cdot,h}+v_{\cdot,\cdot}.
\end{equation}
Here, dots denote equal-weight averaging over the corresponding dimensions.
Positive values indicate higher VCR than the additive baseline predicts.

\paragraph{Resource metrics.}

Token Load is mean token consumption over all scheduled attempts, including
failures. Tokens per Clean Run and Tokens per Success divide total consumption
by the respective counts. Comparisons require 100\% token coverage in matched
model--benchmark blocks. Token-use wins favor lower load; outcome-preserving
wins require no higher load and no lower official outcome, with at least one
strict inequality. Pareto-frontier share measures the fraction of blocks where a
harness is undominated. Relative load summarizes token demand within matched
model--benchmark blocks; its aggregation is detailed in
\Cref{app:scoring-aggregation}. Tokens quantify logical volume
(\Cref{app:detailed-resource-accounting}).

\paragraph{Uncertainty and sensitivity.}

We use 2,000 paired task-cluster bootstrap replicates, stratified by benchmark
and by environment within VitaBench. Sampled tasks retain all repetitions and
model--harness observations. Each replicate recomputes outcome, operational
validity, interaction, Fit, and matched resource statistics, including each
Fit block's best-harness reference and Token Loads from provider-token records.
We report pointwise percentile 95\% intervals without multiplicity adjustment;
cell-level resource summaries remain descriptive. Intervals summarize task-level sampling variation within the evaluated benchmark suite.
Leave-one-benchmark-out analyses assess dependence on benchmark composition.

\section{Experimental Setup}
\label{sec:experiment-setup}

We evaluate model--harness pairings on matched tasks using the metrics in
\Cref{sec:metric-design}. Implementation and evaluator details are provided in
\Cref{app:benchmark-setup}.

\subsection{Benchmark Suite}
\label{sec:benchmark-selection}

We select benchmarks requiring closed-loop action in an external environment
or persistent artifact, matched replay from prescribed states with official
evaluators, and complementary harness-sensitive demands.
\Cref{tab:benchmark-coverage} summarizes their task counts and interaction demands.

\begin{table}[!htbp]
\centering
\caption{Benchmark coverage and harness-sensitive interaction demands.}
\label{tab:benchmark-coverage}
\footnotesize
\setlength{\tabcolsep}{3pt}
\renewcommand{\arraystretch}{1.0}
\resizebox{\textwidth}{!}{%
\begin{tabular}{@{}lrl@{}}
\toprule
\textbf{Benchmark} & \textbf{Tasks} & \textbf{Key interaction demands} \\
\midrule
\textbf{SWE-bench-45} {\citep{jimenez2024swebench,openai2024swebenchverified}}
& 45 & Repository editing, testing, patch handoff \\
\textbf{Terminal-Bench 2} {\citep{merrill2026terminalbench}}
& 89 & Stateful terminal interaction \\
\textbf{VitaBench} {\citep{he2026vitabench}}
& 400 & Multi-turn services, domain APIs \\
\textbf{LOCA-Bench} {\citep{zeng2026locabench}}
& 75 & Long-context planning, exact-state completion \\
\textbf{ClawMark} {\citep{meng2026clawmark}}
& 100 & Multi-day workflows, state changes, multimodal inputs \\
\textbf{AgentSkillOS} {\citep{li2026agentskillos}}
& 30 & Skill selection, artifact handoff \\
\bottomrule
\end{tabular}%
}
\end{table}

\subsection{Models and Harnesses}
\label{sec:models-and-harnesses}

We evaluate four general-purpose harnesses: DeepAgents
\citep{langchain2026deepagents}, OpenCode \citep{opencode2026software}, Pi CLI
\citep{pi2026software}, and Codex CLI \citep{openai2025codexcli}. The core
models are DeepSeek V4 Flash Preview \citep{deepseekai2026v4flash}, MiniMax M3
\citep{minimax2026m3}, Kimi K2.7 \citep{moonshotai2026kimi27}, and
Qwen3.6-35B-A3B \citep{qwen2026qwen36}.
Two fixed native references complement the crossed panels: GPT-5.5 xhigh
\citep{openai2026gpt55} with Codex CLI, and DeepSeek V4 Pro 0813
\citep{deepseekai2026v4pro} with DeepSeek Harness
\citep{deepseekai2026harness}.

\subsection{Evaluation Protocol}
\label{sec:agent-adapter}
\label{sec:crossed-protocol}

\paragraph{Matched execution.}
Our common agent adapter initializes tasks and connects models through each
harness's native interface. It matches tool permissions, resource budgets,
and termination conditions while preserving native prompts, observation and
action interfaces, context management, and execution loops, including recovery
and stopping. Original benchmark evaluators score
results under unchanged task semantics and success criteria.

\paragraph{Evaluation panels.}
Within each benchmark, all supported configurations run the same tasks with
$R=4$ repetitions; comparisons fix the task and repetition while varying one
component. Four benchmarks use full $4\times4$ model--harness panels;
ClawMark and AgentSkillOS use $3\times4$ panels because DeepSeek V4 Flash
Preview lacks the required multimodal support. Cross-benchmark summaries use
the common three-model, four-harness panel of 35,472 scheduled attempts in
72 configurations. The primary AgentSkillOS condition is used in this panel;
fixed references and stress conditions are reported separately. Fixed native systems provide matched complete-system references outside the
crossed indices (\Cref{app:native-system-reference}). Across all core panels,
fixed references, and the AgentSkillOS stress condition, we analyze
\PaperTrajectoryCount{} trajectories.

\paragraph{Scoring and aggregation.}
The primary outcome is the mean official task score over four repetitions,
with binary success or continuous credit as appropriate for each benchmark.
Operational validity requires normal termination and a valid evaluator
handoff, irrespective of task correctness. Benchmark-specific scoring and
cross-benchmark aggregation are given in
\Cref{app:benchmark-metrics,app:scoring-aggregation}.

\Needspace{12\baselineskip}
\section{Results and Implications}
\label{sec:results}

To understand agent failures, we examine task achievement, valid completion,
and resource use across model--harness pairings. We address three research questions:
\begin{itemize}[leftmargin=0pt,label={},topsep=3pt,itemsep=0pt,parsep=0pt]
  \item \textbf{RQ1: Model--Harness Interaction.} How do task performance,
  valid completion, and resource use vary across model--harness pairings and
  benchmarks?
  \item \textbf{RQ2: Cross-Model Fit.} How do harnesses compare across models
  when task performance and valid completion are evaluated jointly?
  \item \textbf{RQ3: Metric Validity and Robustness.} What information do the
  proposed metrics add beyond task scores, and how sensitive are their
  conclusions to aggregation choices and the evaluated models and benchmarks?
\end{itemize}

\begin{table*}[!t]
\centering
\scriptsize
\setlength{\tabcolsep}{5pt}
\renewcommand{\arraystretch}{1.0}
\caption{Main outcomes (0--100; scoring in \Cref{app:benchmark-metrics}).
Best/Mean: Best-of-4/Mean-of-4; ClawMark: continuous scores; AgentSkillOS:
Task Score (\texttt{skill\_200}). Bold/underlined: best/second-best harness per core model and
metric, including ties. Fixed references are excluded from highlighting and
crossed analyses. Dashes: unevaluated; both DeepSeek models lack multimodal
support for ClawMark and AgentSkillOS.}
\label{tab:main-results-summary}
\begin{tabular}{llcc@{\hspace{10pt}}cc@{\hspace{10pt}}cc@{\hspace{10pt}}cc@{\hspace{10pt}}cc@{\hspace{10pt}}c}
\toprule
\multirow{2}{*}{Model} & \multirow{2}{*}{Harness}
 & \multicolumn{2}{c}{\makecell{SWE-\\bench-45}}
 & \multicolumn{2}{c}{\makecell{Terminal-\\Bench 2}}
 & \multicolumn{2}{c}{VitaBench} & \multicolumn{2}{c}{LOCA-Bench}
 & \multicolumn{2}{c}{ClawMark}
 & \multirow{2}{*}{\makecell{AgentSkill\\OS}} \\
\cmidrule(lr){3-4}\cmidrule(lr){5-6}\cmidrule(lr){7-8}
\cmidrule(lr){9-10}\cmidrule(lr){11-12}
 & & Best & Mean & Best & Mean &
 Best & Mean & Best & Mean & Best & Mean & \\
\midrule
\multirow{4}{*}{\makecell[l]{DeepSeek V4\\Flash Preview}}
 & DeepAgents & 40.00 & 31.11 & \textbf{61.80} & \textbf{45.79} & 69.00 & 38.00 & 40.00 & 20.67 & -- & -- & -- \\
 & OpenCode   & \underline{42.22} & 32.22 & \underline{53.93} & 40.45 & 71.50 & 45.38 & \underline{53.33} & \underline{31.00} & -- & -- & -- \\
 & Pi CLI     & \textbf{46.67} & \underline{35.56} & \textbf{61.80} & \underline{44.66} & \underline{75.75} & \underline{48.94} & 42.67 & 29.00 & -- & -- & -- \\
 & Codex CLI  & \textbf{46.67} & \textbf{36.11} & 50.56 & 37.64 & \textbf{77.00} & \textbf{49.06} & \textbf{76.00} & \textbf{50.67} & -- & -- & -- \\
\midrule
\multirow{4}{*}{MiniMax M3}
 & DeepAgents & 55.56 & 31.11 & 46.07 & 21.63 & \underline{77.25} & \textbf{51.00} & \underline{33.33} & \underline{24.00} & 75.06 & 54.78 & 80.07 \\
 & OpenCode   & \textbf{82.22} & \textbf{62.78} & \textbf{55.06} & \textbf{39.61} & 75.50 & \underline{49.38} & 32.00 & 21.00 & \textbf{82.39} & \textbf{69.62} & \textbf{93.49} \\
 & Pi CLI     & \underline{71.11} & \underline{52.22} & \underline{53.93} & 33.15 & \underline{77.25} & 49.25 & 21.33 & 12.67 & \underline{78.34} & \underline{62.58} & \underline{85.39} \\
 & Codex CLI  & 55.56 & 30.56 & 51.69 & \underline{34.55} & \textbf{77.50} & 45.06 & \textbf{93.33} & \textbf{87.00} & 77.25 & 58.49 & 36.44 \\
\midrule
\multirow{4}{*}{Kimi K2.7}
 & DeepAgents & 31.11 & 15.56 & 44.94 & 30.62 & 66.00 & \underline{43.06} & 17.33 & 8.33 & \underline{72.56} & \underline{58.65} & \underline{89.92} \\
 & OpenCode   & \textbf{66.67} & \textbf{48.89} & \underline{53.93} & \underline{39.04} & 68.50 & \textbf{44.38} & 42.67 & 27.67 & \textbf{81.83} & \textbf{70.36} & 84.94 \\
 & Pi CLI     & \underline{51.11} & \underline{42.22} & \textbf{59.55} & \textbf{43.82} & \underline{68.75} & 40.25 & \underline{50.67} & \underline{34.33} & 12.08 & 11.93 & \textbf{95.91} \\
 & Codex CLI  & 28.89 & 10.56 & 48.31 & 31.74 & \textbf{72.00} & 41.44 & \textbf{86.67} & \textbf{54.00} & 12.20 & 11.96 & 55.66 \\
\midrule
\multirow{4}{*}{\makecell[l]{Qwen3.6-\\35B-A3B}}
 & DeepAgents & 31.11 & 15.00 & 40.45 & \underline{26.40} & \underline{59.25} & 27.69 & 14.67 & 8.67 & 68.58 & 52.71 & \underline{74.01} \\
 & OpenCode   & \underline{42.22} & 22.22 & 40.45 & 26.12 & 58.75 & \underline{30.88} & \underline{34.67} & \underline{21.00} & \underline{76.25} & \textbf{63.62} & \textbf{76.12} \\
 & Pi CLI     & 33.33 & \underline{25.00} & \underline{43.82} & \textbf{28.65} & \textbf{59.50} & \textbf{31.19} & 28.00 & 15.00 & \textbf{76.91} & 52.61 & 56.23 \\
 & Codex CLI  & \textbf{44.44} & \textbf{26.67} & \textbf{47.19} & \textbf{28.65} & 51.75 & 22.88 & \textbf{70.67} & \textbf{36.33} & 75.19 & \underline{61.21} & 63.06 \\
\midrule
\multicolumn{2}{l}{GPT-5.5 xhigh + Codex CLI}
 & 86.67 & 65.56 & 76.40 & 60.39 & 69.00 & 46.56 & 90.67 & 76.00 & 82.54 & 74.31 & 97.57 \\
\midrule
\multicolumn{2}{l}{\makecell[l]{DeepSeek V4 Pro 0813\\+ DeepSeek Harness}}
 & 88.89 & 70.56 & 62.92 & 46.91 & 79.25 & 54.13 & 90.67 & 73.33 & -- & -- & -- \\
\bottomrule
\end{tabular}
\end{table*}

\subsection{Model--Harness Interaction}
\label{sec:results-findings}
\label{sec:results-task-performance}
\label{sec:results-crossbench-operability}
\label{sec:results-pairing-affinity}

\begin{figure}[!htb]
\centering
\setlength{\abovecaptionskip}{4pt}
\includegraphics[width=\textwidth]{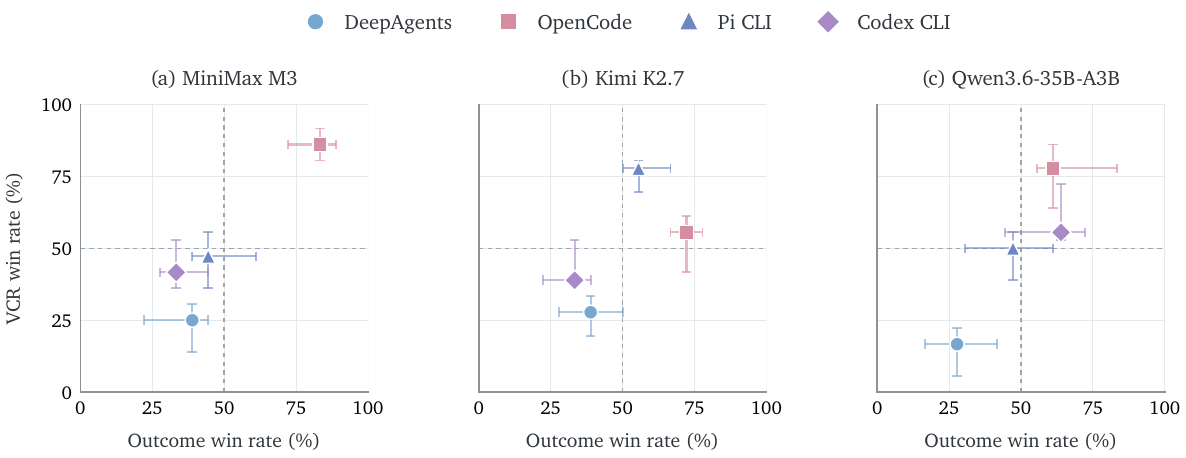}
\caption{Task performance and operational validity on the balanced
six-benchmark panel, summarized by Outcome Win Rate and VCR Win Rate.
Each point represents a model--harness pairing;
horizontal and vertical whiskers give the respective pointwise 95\%
task-cluster bootstrap intervals. Dashed lines: 50\% (neutral).}
\label{fig:performance-operability-slope}
\end{figure}

\paragraph{Task performance and operational validity favor different pairings.}
Harness changes can reverse model rankings within a benchmark (RQ1). On
SWE-bench-45, DeepSeek V4 Flash Preview outperforms MiniMax with Codex CLI,
while MiniMax leads with OpenCode (\Cref{tab:main-results-summary}). Harness
preferences also depend on the metric: for Kimi, OpenCode leads Outcome Win
Rate and Pi CLI leads VCR Win Rate; for Qwen, Codex CLI and OpenCode lead,
respectively (\Cref{fig:performance-operability-slope}). On LOCA-Bench,
MiniMax--Codex also outperforms both native references
(\Cref{app:native-system-reference}).

\paragraph{Consistent valid completion across harnesses.}
DeepSeek V4 Flash Preview maintains high Mean VCR across all four harnesses
on SWE-bench-45, Terminal-Bench 2, VitaBench, and LOCA-Bench:
91.30\%--93.98\%, a range of 2.68 percentage points. Task outcomes still
depend on the harness and benchmark
(\Cref{tab:main-results-summary}).

\begin{figure}[!htb]
\centering
\setlength{\abovecaptionskip}{4pt}
\includegraphics[width=\textwidth]{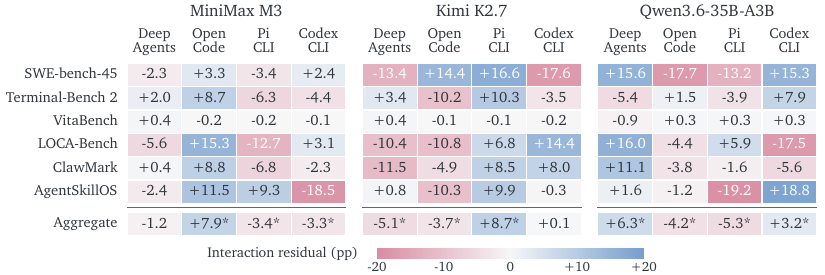}
\caption{Additive VCR interaction residuals (pp) for the balanced
three-model panel. $*$ marks pointwise 95\% task-cluster bootstrap intervals
excluding zero. Benchmark cells report
benchmark-specific point estimates.}
\label{fig:crossbench-pairing-affinity}
\end{figure}

\paragraph{Pairing-specific deviations complement component averages.}
MiniMax--OpenCode and Kimi--Pi exceed additive VCR baselines, with positive
aggregate residuals whose pointwise bootstrap intervals exclude zero
(\Cref{fig:crossbench-pairing-affinity}). Deviations vary by benchmark:
Kimi--Codex is below baseline on SWE-bench-45 but above it on LOCA-Bench.
Opposing deviations can cancel in aggregate, so benchmark-level profiles
reveal interactions hidden by aggregate summaries.

\begin{figure}[!htb]
\centering
\setlength{\abovecaptionskip}{4pt}
\includegraphics[width=\textwidth]{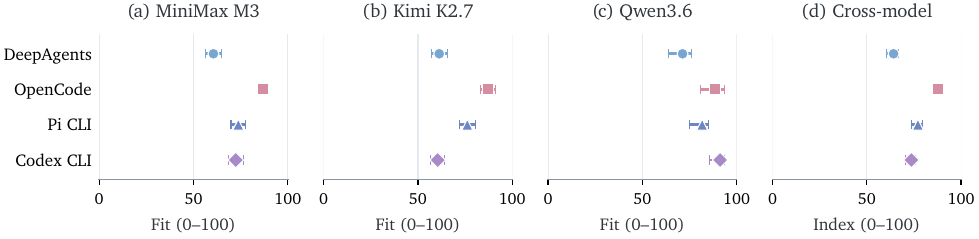}
\caption{Model--harness Fit and Cross-Model Fit on the balanced six-benchmark
panel (0--100). Whiskers: pointwise 95\% paired task-cluster bootstrap
intervals. Higher values retain more of the best observed joint score.}
\label{fig:cross-model-fit}
\end{figure}

\paragraph{Cross-model fit complements model-specific choice.}
OpenCode has the highest Cross-Model Fit and similar Fit across all three
models (RQ2; \Cref{fig:cross-model-fit}). Codex CLI leads for Qwen, but its
lower Kimi Fit reduces its aggregate. Each Fit is relative to that model's
best observed harness on each benchmark: the aggregate supports a shared
harness choice, while model-specific profiles guide selection for a known model.

% Generated by scripts/generate_fit_results.py.
\begin{table}[!htb]
\centering
\small
\setlength{\tabcolsep}{5pt}
\caption{Fit on the balanced six-benchmark panel (0--100). Brackets give pointwise 95\% paired task-cluster bootstrap intervals.}
\label{tab:fit-model-details}
\begin{tabular}{lcccc}
\toprule
Harness & MiniMax M3 & Kimi K2.7 & Qwen3.6 & Cross-model \\
\midrule
OpenCode & \makecell{86.86\\{\scriptsize [85.11, 88.50]}} & \makecell{87.06\\{\scriptsize [83.02, 90.94]}} & \makecell{88.55\\{\scriptsize [80.84, 93.36]}} & \makecell{87.48\\{\scriptsize [84.06, 89.99]}} \\
Pi CLI & \makecell{73.73\\{\scriptsize [69.87, 77.73]}} & \makecell{76.16\\{\scriptsize [71.99, 80.27]}} & \makecell{81.70\\{\scriptsize [74.88, 84.98]}} & \makecell{77.13\\{\scriptsize [73.49, 79.62]}} \\
Codex CLI & \makecell{72.37\\{\scriptsize [68.59, 76.61]}} & \makecell{60.45\\{\scriptsize [56.84, 63.91]}} & \makecell{91.21\\{\scriptsize [85.55, 93.44]}} & \makecell{73.62\\{\scriptsize [70.54, 75.82]}} \\
DeepAgents & \makecell{60.56\\{\scriptsize [56.35, 64.62]}} & \makecell{61.29\\{\scriptsize [57.11, 65.48]}} & \makecell{71.29\\{\scriptsize [63.99, 76.22]}} & \makecell{64.20\\{\scriptsize [60.57, 66.94]}} \\
\bottomrule
\end{tabular}
\end{table}

\subsection{Joint Outcomes and Execution Diagnostics}
\label{app:additional-execution}
\label{app:joint-pairing-profiles}
\label{sec:results-operability-mechanisms}

\begin{figure}[!htb]
\centering
\setlength{\abovecaptionskip}{4pt}
\includegraphics[width=\textwidth]{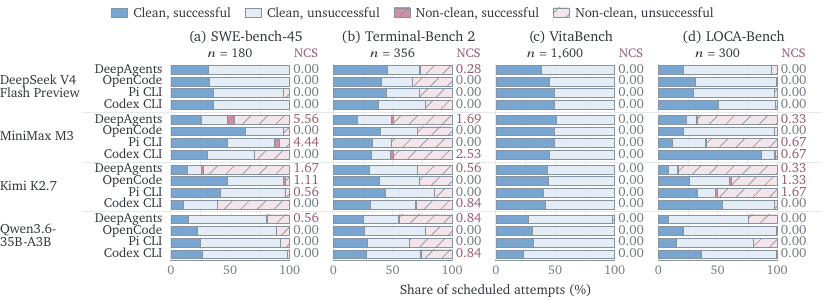}
\caption{Joint outcome--validity shares for all 64 configurations
($n$: attempts per pairing). Bars sum to 100\%; NCS gives non-clean success
(\%). Four repetitions per task; fixed references are excluded.}
\label{fig:joint-outcome-operability}
\end{figure}

\paragraph{Task achievement and valid completion can diverge.}
Non-clean attempts can earn task credit, while clean attempts can leave a task
unsolved (\Cref{fig:joint-outcome-operability,fig:joint-score-contributions}).
We examine this divergence through continuous-score contributions, execution
stages, and a ClawMark action-realization audit.

\subsubsection{Continuous-Score Decomposition}
\label{app:joint-continuous}

\begin{figure}[!htb]
\centering
\includegraphics[width=\textwidth]{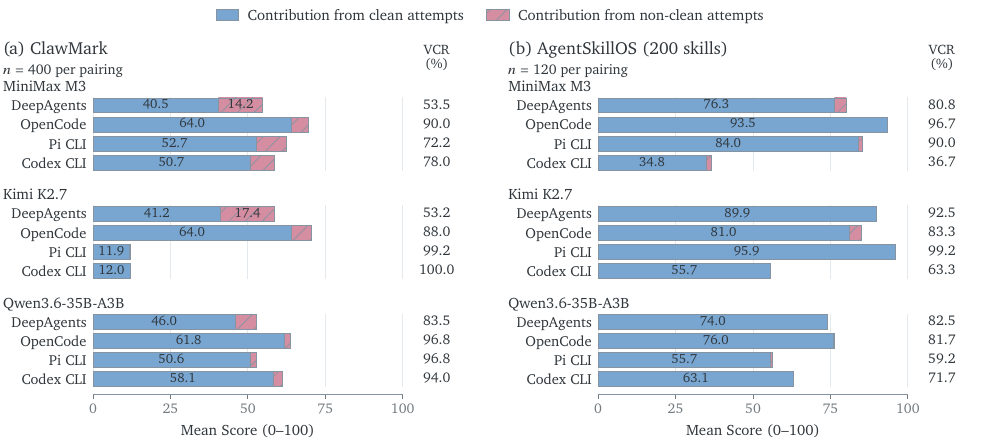}
\caption{Clean/non-clean contributions summing to Mean Score. Right column:
VCR (\%).
ClawMark: 400 attempts per configuration; AgentSkillOS (\texttt{skill\_200}):
120. Four repetitions per task; fixed references are excluded.}
\label{fig:joint-score-contributions}
\end{figure}

For ClawMark and AgentSkillOS (\texttt{skill\_200}), we decompose Mean Score
into clean and non-clean contributions. Let $z_i$ be attempt $i$'s official
score on a 0--100 scale, $U_i$ its clean-run indicator, and $n$ the number
of scheduled attempts:
\begin{equation}
\mathrm{Mean\ Score}
=\frac{1}{n}\Bigl(
\underbrace{\sum_{i:U_i=1} z_i}_{\text{clean}}
+\underbrace{\sum_{i:U_i=0} z_i}_{\text{non-clean}}
\Bigr).
\end{equation}
The clean and non-clean contributions sum to the original Mean Score
(\Cref{fig:joint-score-contributions}).

On ClawMark, Kimi--DeepAgents exceeds the fully clean Kimi--Codex by 46.7
Mean Score points, with nearly a third of its score coming from non-clean
attempts. On AgentSkillOS, MiniMax--Codex's mean within clean attempts exceeds
its overall Task Score by 58.4 points, but clean attempts cover just over a
third of scheduled runs.

\subsubsection{Execution-Stage Decomposition}
\label{app:execution-stage-decomposition}

\begin{figure}[!htb]
\centering
\includegraphics[width=\textwidth]{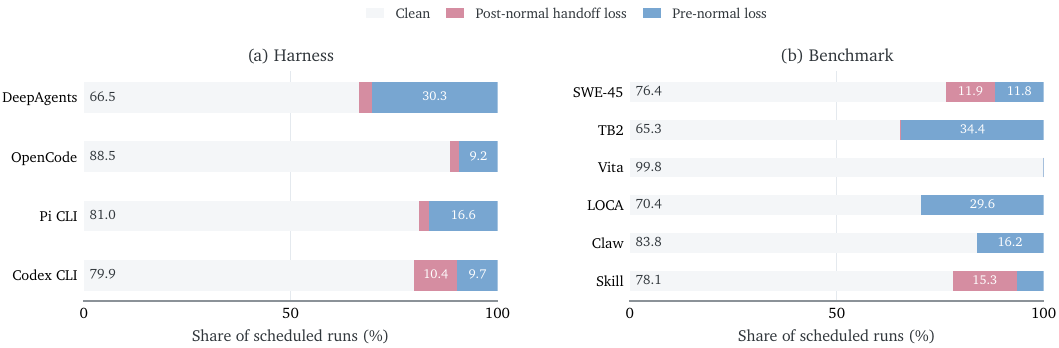}
\caption{Exclusive execution stages on the balanced three-model panel: clean
completion, pre-normal loss, and post-normal handoff loss sum to 100\%.
Panels report means by (a) harness and (b) benchmark. Fixed references are excluded.}
\label{fig:failure-stage-decomposition}
\end{figure}

We partition non-clean runs into pre-normal loss (abnormal termination) and
post-normal handoff loss (normal termination without a valid evaluator state).
These exclusive stages and VCR sum to 100\%
(\Cref{fig:failure-stage-decomposition}).

DeepAgents loses nearly ten times as many runs before normal
termination as at post-normal handoff. OpenCode and Pi CLI show the same
pattern at lower rates; Codex CLI's losses are roughly balanced between the
two stages.

Terminal-Bench 2, LOCA-Bench, and ClawMark mainly expose pre-normal failures,
whereas VitaBench is near ceiling. SWE-bench-45 and AgentSkillOS also expose
handoff losses; AgentSkillOS loses 3.4 percentage points more attempts at this
stage than SWE-bench-45.

Pre-normal losses direct diagnosis toward interface adherence, timeouts, and
recovery; handoff losses toward stopping, finalization, and submission
validation. Low scores after clean completion direct trajectory inspection
toward planning, action realization, and task-specific support.

\subsubsection{Case Study: Action Realization and Clean Completion}
\label{sec:clawmark-action-realization}

We audit Kimi K2.7 on ClawMark to examine whether stated action intent becomes
executable actions. Codex CLI and Pi CLI achieve 100.00\% and 99.25\% VCR,
yet score only 11.96 and 11.93. OpenCode scores 58.4 points above Codex CLI
with 88.00\% VCR; DeepAgents scores 58.65 with 53.25\% VCR
(\Cref{tab:clawmark-action-realization}).

A scripted indicator identifies strict \emph{action-realization gaps}: prospective action
intent in reasoning with neither a structured action nor non-whitespace
visible text in the same accepted response. Provider retries and no-tool
context-compaction generations are treated separately. Native records and
control flow classify all 400 attempts per harness; for 48 timed-out OpenCode
attempts, its no-fallback control flow establishes no strict gaps in
unrecorded stages.

\begin{table}[htbp]
\centering
\small
\setlength{\tabcolsep}{5pt}
\caption{Strict Kimi K2.7 action-realization audit on ClawMark. Each harness has
400 attempts. Brackets show pointwise 95\% task-cluster bootstrap intervals.
For no-event cells, [0,0] reflects zero empirical bootstrap variation within
the observed sample. Dashes denote undefined ratios.}
\label{tab:clawmark-action-realization}
\begin{tabular}{lrrcc}
\toprule
Harness & \makecell{Score\\Mean} & VCR &
\makecell{Affected attempts\\\% [95\% CI]} &
\makecell{Events per\\affected attempt} \\
\midrule
Codex CLI  & 11.96 & 100.00 &  6.75 [4.50, 9.25]   & 1.00 \\
DeepAgents & 58.65 &  53.25 &  0.00 [0.00, 0.00]   & --   \\
OpenCode   & 70.36 &  88.00 &  0.00 [0.00, 0.00]   & --   \\
Pi CLI     & 11.93 &  99.25 & 40.50 [36.25, 44.75] & 1.32 \\
\bottomrule
\end{tabular}
\end{table}

Strict gaps affect 6.75\% of Codex CLI attempts and 40.50\% of Pi CLI attempts,
with none in OpenCode or DeepAgents. Pi has six times as many affected
attempts and recurring events; Codex has one event per affected attempt.
This pattern coincides with the score split across harnesses, while affected
attempts score less than 1.3 points below unaffected attempts within each
of Codex and Pi.

Recovery policies differ: Codex retries malformed-action JSON, Pi restarts only
after continuation-role errors, OpenCode has no external fallback, and
DeepAgents runs a finalization pass. All but one Codex event and every Pi event
have no linked later response; the remaining Codex retry is unresolved.
High VCR can thus coexist with action-realization gaps. Trajectory-level
diagnostics therefore complement VCR.

\subsection{Metric Validation and Robustness}
\label{sec:metric-validation}
\label{app:fit-validation}

For RQ3, we compare Fit components and assess ranking sensitivity to
sampling, normalization references, and panel composition.

% Generated by scripts/generate_fit_results.py.
\begin{table}[!t]
\centering
\small
\setlength{\tabcolsep}{5pt}
\renewcommand{\arraystretch}{1.0}
\caption{Comparison metrics and Fit ablations on the common three-model panel. Outcome-only replaces $US$ with $S$; arithmetic replaces the cross-model geometric mean. Aggregation details: \Cref{app:scoring-aggregation}.}
\label{tab:fit-ablation}
\begin{tabular}{lrrrrr}
\toprule
& \multicolumn{2}{c}{Win rate (\%)} & \multicolumn{3}{c}{Fit variants (0--100)} \\
\cmidrule(lr){2-3}\cmidrule(lr){4-6}
Harness & Outcome & VCR & Outcome-only & Arithmetic & Cross-model \\
\midrule
OpenCode & 72.22 & 73.15 & 87.84 & 87.49 & 87.48 \\
Pi CLI & 49.07 & 58.33 & 78.20 & 77.20 & 77.13 \\
Codex CLI & 43.52 & 45.37 & 74.56 & 74.68 & 73.62 \\
DeepAgents & 35.19 & 23.15 & 68.15 & 64.38 & 64.20 \\
\bottomrule
\end{tabular}
\end{table}

\paragraph{Fit components and diagnostic value.}
The outcome-only variant removes the valid-completion gate; the arithmetic
variant replaces the cross-model geometric mean (\Cref{tab:fit-ablation}).
Joint and official scores coincide in 37 of
72 benchmark-level cells; counting credit only from valid completions
reduces scores in the other 35. For MiniMax on SWE-bench-45, this reverses the DeepAgents--Codex
ordering: DeepAgents scores 31.11 officially and 25.56 jointly, whereas
Codex scores 30.56 under both measures. On ClawMark, non-clean attempts
contribute 17.43 points to Kimi--DeepAgents' score, the largest cell-level
gap.

Geometric aggregation lowers Codex CLI's index by 1.06 points relative to
the arithmetic variant, the largest reduction among the four harnesses.
Both substitutions preserve the overall ordering; the outcome-only variant
also preserves each model's harness ordering.

\paragraph{Uncertainty and panel sensitivity.}
The Pi--Codex Cross-Model Fit difference is 3.50 points, with a pointwise
95\% paired bootstrap interval of $[-0.26,7.05]$
(resampling in \Cref{sec:metric-design}).

% Generated by scripts/generate_fit_results.py.
\begin{table}[!htb]
\centering
\small
\setlength{\tabcolsep}{6pt}
\caption{Cross-model Fit after omitting one model or benchmark. Calculation details: \Cref{app:scoring-aggregation}. DA: DeepAgents; OC: OpenCode; Pi: Pi CLI; CX: Codex CLI.}
\label{tab:fit-sensitivity}
\begin{tabular}{llrrrrl}
\toprule
Omitted & & DA & OC & Pi & CX & Order \\
\midrule
Model & MiniMax & 66.10 & 87.80 & 78.88 & 74.26 & $OC > Pi > CX > DA$ \\
Model & Kimi & 65.71 & 87.70 & 77.61 & 81.25 & $OC > CX > Pi > DA$ \\
Model & Qwen & 60.92 & 86.96 & 74.94 & 66.14 & $OC > Pi > CX > DA$ \\
\midrule
Benchmark & SWE-45 & 68.86 & 86.06 & 75.45 & 77.75 & $OC > CX > Pi > DA$ \\
Benchmark & TB2 & 63.29 & 86.29 & 73.66 & 71.91 & $OC > Pi > CX > DA$ \\
Benchmark & Vita & 57.82 & 85.26 & 73.40 & 70.69 & $OC > Pi > CX > DA$ \\
Benchmark & LOCA & 72.63 & 96.23 & 84.82 & 67.98 & $OC > Pi > DA > CX$ \\
Benchmark & Claw & 63.60 & 84.98 & 80.19 & 76.23 & $OC > Pi > CX > DA$ \\
Benchmark & Skill & 58.87 & 86.01 & 74.88 & 76.59 & $OC > CX > Pi > DA$ \\
\bottomrule
\end{tabular}
\end{table}

Omitting any harness preserves the survivors' relative ordering, but changes
their scores when the omitted harness supplied a best-score reference.
Recomputing Fit after omitting any single model or benchmark leaves OpenCode
first (\Cref{tab:fit-sensitivity}). Intermediate ranks vary: omitting Kimi
places Codex CLI above Pi CLI, while omitting LOCA-Bench places it below
DeepAgents.

\paragraph{Leave-one-model-out selection diagnostic.}
For each held-out model, we select a harness using the two remaining models
and evaluate its Fit against the held-out model's full harness panel.
Outcome Win Rate, VCR Win Rate, outcome-only Fit, and joint Fit all select
OpenCode in all three folds. It is the best harness for MiniMax and Kimi;
its Qwen Fit is 2.66 points below Codex CLI
(\Cref{tab:fit-model-holdout}). This retrospective diagnostic compares
selection across the observed models.

% Generated by scripts/generate_fit_results.py.
\begin{table}[!htb]
\centering
\small
\setlength{\tabcolsep}{5pt}
\caption{Leave-one-model-out selection diagnostic. Fit, outcome-only Fit, Outcome Win Rate, and VCR Win Rate all select OpenCode using the two remaining models. Regret is the difference, in Fit points, between the held-out model's best single-harness Fit and the selected Fit.}
\label{tab:fit-model-holdout}
\begin{tabular}{lclrr}
\toprule
Held-out model & Selected Fit & Best harness & Best Fit & Regret \\
\midrule
MiniMax M3 & 86.86 & OpenCode & 86.86 & 0.00 \\
Kimi K2.7 & 87.06 & OpenCode & 87.06 & 0.00 \\
Qwen3.6 & 88.55 & Codex CLI & 91.21 & 2.66 \\
\bottomrule
\end{tabular}
\end{table}

\subsection{Resource Use and Repeated-Run Behavior}
\label{sec:results-token-efficiency}

\paragraph{Resource use informs harness choice.}
Across complete primary comparison blocks, Pi CLI leads token-use win rate,
while OpenCode leads outcome-preserving win rate and Pareto-frontier share
(\Cref{tab:relative-token-indices}). Codex CLI has the highest outcome and
lowest tokens per success for every model on LOCA-Bench. On ClawMark,
OpenCode outperforms DeepAgents with fewer tokens for all three models,
whereas Kimi's low-load Pi and Codex runs score only 11.93 and 11.96.
Token Load must be read alongside outcomes: low load can reflect early
failure, and high load can include repeated ineffective actions.

% Generated by scripts/generate_resource_results.py; do not edit by hand.
\begin{table}[!ht]
\centering
\small
\setlength{\tabcolsep}{3.5pt}
\vspace{-8pt}
\caption{Matched resource comparisons (18 complete primary model--benchmark
blocks; 72 cells). First three columns: \%; brackets: pointwise 95\% bootstrap
intervals; bold/underlined: best/second-best. Definitions and resampling:
\Cref{sec:metric-design}.}
\label{tab:relative-token-indices}
\begin{tabular}{lcccc}
\toprule
Harness & \makecell{Token-use\\win rate} & \makecell{Outcome-preserving\\win rate} & \makecell{Pareto-frontier\\share} & \makecell{Relative\\load} \\
\midrule
DeepAgents & \makecell{\underline{51.85}\\{\scriptsize [44.44, 53.70]}} & \makecell{14.81\\{\scriptsize [11.11, 22.22]}} & \makecell{50.00\\{\scriptsize [27.78, 55.56]}} & \makecell{1.095\\{\scriptsize [1.043, 1.159]}} \\
OpenCode & \makecell{44.44\\{\scriptsize [40.74, 51.85]}} & \makecell{\textbf{35.19}\\{\scriptsize [31.48, 44.44]}} & \makecell{\textbf{61.11}\\{\scriptsize [50.00, 72.22]}} & \makecell{1.150\\{\scriptsize [1.112, 1.202]}} \\
Pi CLI & \makecell{\textbf{55.56}\\{\scriptsize [50.00, 61.11]}} & \makecell{\underline{24.07}\\{\scriptsize [20.37, 33.33]}} & \makecell{50.00\\{\scriptsize [38.89, 66.67]}} & \makecell{\underline{0.906}\\{\scriptsize [0.827, 0.967]}} \\
Codex CLI & \makecell{48.15\\{\scriptsize [44.44, 51.85]}} & \makecell{\underline{24.07}\\{\scriptsize [16.67, 27.78]}} & \makecell{\underline{55.56}\\{\scriptsize [44.44, 61.11]}} & \makecell{\textbf{0.877}\\{\scriptsize [0.838, 0.927]}} \\
\bottomrule
\end{tabular}
\vspace{-8pt}
\end{table}

\subsubsection{Repeated-Run Outcomes and Stress Conditions}
\label{app:repeatability-stress}

\begin{figure}[!htb]
\centering
\includegraphics[width=\textwidth]{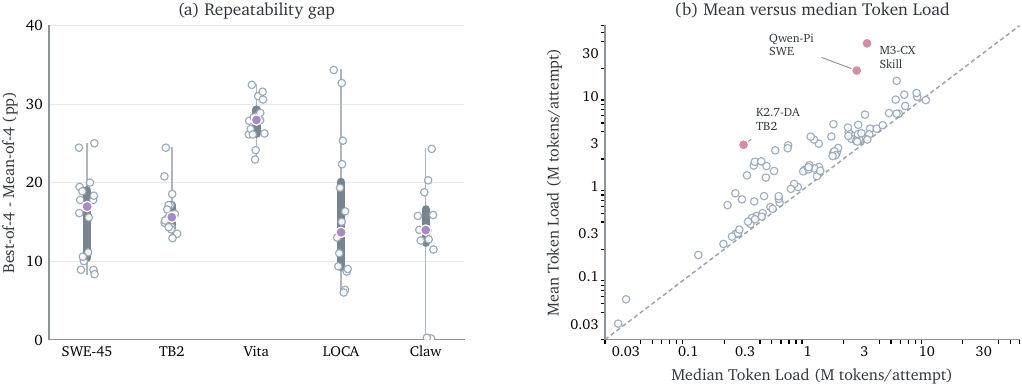}
\caption{Repeated-run gaps and resource tails. (a) Best-of-4 minus Mean-of-4
for 76 core crossed configurations; thick segments: interquartile ranges;
filled points: medians. (b) Mean versus median Token Load for
\ResourceMedianCells{} fully covered primary cells with both statistics;
dashed line: equality. The three most skewed cells are highlighted.}
\label{fig:stability-resource-tails}
\end{figure}

Best-of-4 exceeds Mean-of-4 in all 76 configurations in
\Cref{fig:stability-resource-tails}(a). VitaBench has the largest median gap,
about 14 percentage points above LOCA-Bench's smallest gap.
MiniMax--OpenCode's gap on SWE-bench-45 is 19.4 points. Fixed references
show the same pattern on SWE-bench-45, Terminal-Bench 2, and VitaBench.
Run-to-run variation in task success persists alongside reliable protocol
completion.

Under stress, VitaBench's Cross environment is harder than every
single-domain environment for every configuration. Increasing AgentSkillOS's
skill pool from 200 to 900 reduces GPT-5.5--Codex Task Score more than VCR.

\subsubsection{Resource Profiles and Failure Tails}
\label{app:resource-profiles}

Token Load is strongly right-skewed: means exceed medians in
\ResourceMeanAboveMedianCells{} of \ResourceMedianCells{} fully covered
primary cells, and the three highlighted means are roughly seven to twelve
times their medians (\Cref{fig:stability-resource-tails}(b)). Within a
model--benchmark block, harness loads can differ by more than an order of
magnitude; the lightest harness changes across models and benchmarks.
\Cref{tab:resource-cell-results} reports Token Load, tokens per success,
and coverage for all \ResourceFullRows{} primary, stress-condition, and
fixed-reference cells.

\paragraph{Failure tails and fixed systems.}
The highest-token 5\% of Qwen--Pi SWE runs consume about 83\% of that
configuration's tokens, and failed attempts account for almost all its
token use. Fixed systems also show large benchmark differences:
GPT-5.5--Codex uses about fourteen times as many tokens per attempt on
ClawMark as on VitaBench. DeepSeek V4 Pro 0813--DeepSeek Harness uses over
two orders of magnitude more on LOCA-Bench than on VitaBench, with complete
token coverage in both cells; failed attempts consume over three-quarters
of its Terminal-Bench 2 tokens.

\Needspace{4\baselineskip}
\subsection{Implications for Evaluation, Training, and Harness Design}
\label{sec:implications}

\paragraph{Evaluation.}
Ranking reversals motivate crossed evaluation on matched tasks. Use
model-conditioned outcomes, VCR, and execution diagnostics to distinguish
task failure from execution or handoff failure. Cross-Model Fit supports a
harness shared across models; benchmark profiles, uncertainty, and resource
use reveal pairing differences and their costs.

\paragraph{Model training.}
Native- and mixed-harness RL provide trajectories from diverse execution
interfaces \citep{xu2026polar,song2026clawgym2,wei2026harnessbandit}. Our ranking
reversals motivate pairing the same training tasks with multiple harnesses
to develop transferable action realization, feedback use, recovery, and
stopping. Non-clean successes motivate combining task rewards with execution
validity feedback. Evaluate outcomes, VCR, and Fit on held-out tasks and
unseen harnesses under matched training budgets.

\paragraph{Harness design.}
Use model-specific Fit and VCR profiles to guide interface, recovery, and
context-management choices. Feedback should distinguish rejected from
executed operations, and compression should retain task-relevant state.
Adapt resource limits to the model and task, bound failed retries, and
detect stagnation and repeated actions. Reserve verification and handoff
budgets. Component ablations can assess
which mechanisms improve outcomes.

\FloatBarrier

\section{Conclusion}

We presented a framework for evaluating model--harness interaction through
task performance, valid completion, and resource use. Across six benchmarks, harness rankings depend on the model and task, and similar task
scores can accompany different execution profiles. These findings show that
interpreting agent failures requires examining both components and their
particular pairing. Task failure must be distinguished from execution or
handoff failure: valid completion can leave a task unsolved, while abnormal
termination can coexist with task credit. Model--harness Fit and Cross-Model
Fit summarize task performance achieved through valid completion, while VCR and
trajectory diagnostics provide information on execution failures beyond task scores.
These findings support evaluating model--harness pairs and designing harnesses
for reliable, efficient operation across models. They also motivate treating
harness-use capability as a training objective, with the goal of sustaining
strong task performance and valid completion across diverse execution
interfaces, including harnesses unseen during training.

\FloatBarrier
\bibliographystyle{plainnat}
\bibliography{reference}

%%%%%%%%%%%%%%%%%%%%%%%%%%%%%%%%%%%%%%%%%%%%%%%%%%%%%%%%%%%%

\clearpage
\appendix
\crefalias{section}{appendix}
\crefalias{subsection}{subappendix}
\crefalias{subsubsection}{subsubappendix}

%%%%%%%%%%%%%%%%%%%%%%%%%%%%%%%%%%%%%%%%%%%%%%%%%%%%%%%%%%%%

\section{Experimental Configuration Details}
\label{app:benchmark-setup}

Four repetitions of each task are scheduled for every model--harness
configuration unless specified otherwise. Each harness retains its native
prompts and internal agent policy, with compatibility changes for unattended
execution. This appendix records model serving, execution configurations,
and benchmark-specific evaluator mappings.

\subsection{Model Serving and Software Versions}
DeepSeek V4 Flash Preview was evaluated through its official API. MiniMax M3,
Kimi K2.7, and DeepSeek V4 Pro 0813 were served locally with SGLang on a single
node with eight NVIDIA H200 GPUs. The four crossed harnesses used DeepAgents
Code 0.1.3, OpenCode v1.15.6, Pi CLI 0.79.3, and Codex CLI. The
DeepSeek V4 Pro 0813 fixed reference used DeepSeek Harness SDK 0.1.0rc6. The
GPT-5.5 xhigh--Codex CLI fixed reference was run through an official Codex
subscription.

\subsection{Execution and Benchmark Configurations}

\paragraph{Execution environment and records.}
Runs use fresh task-scoped sandboxes without elevated privileges,
isolated from evaluator internals, shared experiment storage, and other runs;
protected credentials and resource controls remain outside. The adapter records
final submissions or environment states, trajectories, resource use, and
termination status in a common format.

\paragraph{SWE-bench-45.}
SWE-bench Verified assigns each task a human-annotated difficulty label based on
the time an experienced developer is expected to need to identify and implement
a solution. We include every task whose estimate is at least one hour, yielding
45 of the 500 verified tasks: 42 labeled \texttt{1--4 hours} and three labeled
\texttt{>4 hours}. The selected tasks span eight Python repositories, with 22
from Django, seven from SymPy, five from Sphinx, and 11 from the remaining five
repositories. The subset includes all tasks with released difficulty estimates
of at least one hour; ``long'' denotes estimated human completion time.

Each run starts from the prescribed base commit in the official containerized
environment and receives the original GitHub issue description. We retain the
task-specific regression tests and apply the official verifier to the final
repository state. All harnesses operate on the same task images through the
common adapter.

\paragraph{Terminal-Bench 2.}
We execute all 89 tasks through Harbor while retaining their original task
definitions, containerized environments, verifier scripts, and per-task time
limits. Every configuration starts from the same task environment and is
assessed by the same task-specific verifier.

All 16 core configurations contain four distinct trajectories for each of the
89 tasks; both fixed system references use the same $89\times4$ support.

\paragraph{VitaBench.}
We use English task variants for 400 tasks across four environments (100 per
environment), covering multi-turn services, domain-specific APIs, and
cross-domain requests. Each task is repeated four times, giving 400 attempts
per environment and model--harness configuration. A
trajectory is limited to 300 interaction steps. We report the strict
tool-permission setting: harness-native tools and external filesystem, shell,
browsing, and delegation capabilities are disabled where applicable, so the
agent can affect the simulated environment only through the VitaBench business
APIs. DeepSeek-V3.2 serves as both the user simulator and trajectory evaluator.
The simulator reveals task constraints through multi-turn interaction, and the
evaluator applies the binary success criterion in
\Cref{app:benchmark-metrics}.

\paragraph{LOCA-Bench.}
We evaluate all 75 tasks in the 256K-context setting, exercising long-context
state tracking and multi-tool plans. Each task specifies an
initial application state and an exact target state to be reached through the
available tools; the official evaluator compares the final state with that
target.

\paragraph{ClawMark.}
We evaluate 100 tasks across 13 professional scenarios, covering multi-day
workplace activity, silent state changes, and multimodal inputs.
We apply a uniform 30-minute execution limit to every turn and every
model--harness configuration. When a turn reaches this limit, we terminate the
agent, evaluate the service state produced up to that point, and advance the
task to the next in-universe working day. This preserves the benchmark's
continuous workflow while applying the same per-turn budget to every
configuration.

\paragraph{AgentSkillOS.}
We evaluate 30 tasks across five artifact categories, requiring skill
selection, artifact production, and output-contract handoff.
We preserve the original task prompts. The tested context budget accommodates
the released 200-skill primary setting and a 900-skill stress setting containing
all 19 task-required skills and 881 distractors drawn from the released skill
pool. Each task has a 3,600-second wall-clock limit. Streaming harnesses
additionally use a 900-second idle-timeout guard and the harness's retry policy
where supported. Generated artifacts are scored with the official AgentSkillOS
evaluator.

\subsection{Benchmark-Specific Evaluation Metrics}
\label{app:benchmark-metrics}

This section specifies how each benchmark's official evaluator output is
converted into task-level scores. Best-of-4 and
stress-condition results are secondary analyses.

\paragraph{Operational validity status mapping.}
For SWE-bench-45, the normal-termination condition is a recognized normal stop
(\texttt{Submitted}, \texttt{NoPatch}, or a successful exit-zero completion),
and the valid-handoff condition requires the evaluator to accept the patch at
its application stage. For Terminal-Bench 2, normal termination requires a
clean return without a timeout or unhandled framework failure, and valid
handoff requires the post-run verifier to produce a consumable result. For
LOCA-Bench, normal termination requires \texttt{status=success}; valid handoff
requires a finite final-state evaluator
result, including one produced before a subsequently abnormal backend exit.
For ClawMark, normal termination requires \texttt{status=ok}, return code zero, and
no recorded framework failure; valid handoff requires a finite official score.
For AgentSkillOS, normal termination
requires the task runner to record \texttt{status=completed}; valid handoff
requires a newly copied generated file whose path matches, or is nested under,
at least one task-declared expected output path. Task correctness is not
required. For VitaBench, normal termination requires
\texttt{termination\_reason=agent\_stop}, and valid handoff requires a finite
evaluator reward. In every benchmark, correctness affects the
official task outcome but is never part of the valid-handoff condition.

\paragraph{Shared binary aggregation.}
SWE-bench-45, Terminal-Bench 2, and LOCA-Bench have binary run-level outcomes.
For any one of these benchmarks, let $N$ be its number of tasks and let
$a_{i,r}\in\{0,1\}$ denote whether task $i$ succeeds on repetition
$r\in\{1,2,3,4\}$. We report
\begin{align}
\operatorname{Best\text{-}of\text{-}4}
  &= \frac{100}{N}\sum_{i=1}^{N}
     \mathbb{I}\!\left[\max_{1\leq r\leq4}a_{i,r}=1\right], \\
\operatorname{Mean\text{-}of\text{-}4}
  &= \frac{100}{4N}\sum_{i=1}^{N}\sum_{r=1}^{4}a_{i,r}.
\end{align}
Best-of-4 is the fraction of tasks solved at least once in four attempts,
whereas Mean-of-4 is the success probability of a uniformly selected run.

\paragraph{SWE-bench-45.}
Each run starts from the task's prescribed repository state and is evaluated
with the official task-specific regression tests. We set $a_{i,r}=1$ only
when the submitted patch fixes the designated failing tests without breaking
the tests that previously passed. The shared binary aggregation above is
applied with $N=45$.

\paragraph{Terminal-Bench 2.}
The official task-specific verifier evaluates the resulting environment state,
and a run succeeds only when it receives full credit. We apply the shared
binary aggregation with $N=89$.

\paragraph{LOCA-Bench.}
The final-state evaluator returns one when the resulting application state
matches the task's exact target state and zero otherwise. We apply the shared
binary aggregation with $N=75$. A correct evaluator result remains an official task
success even if the backend later records an abnormal exit; such a run is not
a clean run because it fails the normal-termination gate.

\paragraph{VitaBench.}
VitaBench measures complete task success rather than partial progress. Its
sliding-window trajectory evaluator deduplicates the expected-state and overall
rubrics and carries their satisfaction states across overlapping 10-turn
windows with a two-turn overlap. For trajectory $i$, let $\mathcal{Q}_i$ be the
resulting rubric set. Its binary reward is
$R_i=\mathbb{1}[\bigwedge_{q\in\mathcal{Q}_i}q]$: the run receives one only if
every rubric is satisfied.

For each environment $d$, we evaluate $N_d=100$ tasks over four trials. Let
$a_{d,j,r}\in\{0,1\}$ denote the binary reward for task $j$ in trial $r$. We
report
\begin{align}
\operatorname{Best\text{-}of\text{-}4}_d
  &= \frac{100}{N_d}\sum_{j=1}^{N_d}
     \mathbb{I}\!\left[\max_{1\leq r\leq4}a_{d,j,r}=1\right], \\
\operatorname{Mean\text{-}of\text{-}4}_d
  &= \frac{100}{4N_d}\sum_{j=1}^{N_d}\sum_{r=1}^{4}a_{d,j,r}.
\end{align}
For either metric $M$, its Total is the unweighted macro-average over Delivery,
In-store, OTA, and Cross-domain,
$M_{\mathrm{Total}}=\frac{1}{4}\sum_{d\in\mathcal{D}}M_d$.

\paragraph{ClawMark.}
Each task has a weighted rubric implemented by deterministic Python functions
that inspect the resulting state of the sandboxed services. The task score is
the weight-normalized credit across rubric items and lies in $[0,1]$. For each
task, Best-of-4 is the maximum continuous score across four runs and Mean-of-4
is their average. We then average both task-level quantities over all 100 tasks
and multiply by 100 to report scores on a 0--100 scale. Unlike the binary
benchmarks, ClawMark's Best-of-4 retains partial-credit differences below full
success.

\paragraph{AgentSkillOS.}
Task Score is the official AgentSkillOS evaluator score on its native 0--100
scale. We average it over four repetitions of the same 30 tasks. The cross-benchmark
primary outcome uses the \texttt{skill\_200} condition; the 900-skill condition
is reported as a stress test. Execution status is used only for operational validity,
not Task Score.

\subsection{Scoring and Aggregation Details}
\label{app:scoring-aggregation}

\paragraph{Score scales and averaging.}
The primary outcome is the mean official task score over four repetitions:
binary success for SWE-bench-45, Terminal-Bench 2, VitaBench, and LOCA-Bench,
and continuous credit for ClawMark and AgentSkillOS. VitaBench weights its
four environments equally; AgentSkillOS uses the 200-skill primary condition.
All scheduled attempts enter the averages, and benchmarks and models receive
equal weight. Before forming $US$, AgentSkillOS scores are divided by 100;
the other benchmarks retain their native 0--1 scales. Best-of-4 and the
900-skill stress condition are secondary analyses. Benchmark-specific
reporting scales are specified in \Cref{app:benchmark-metrics}.

\paragraph{Win-rate aggregation.}
Outcome Win Rate and VCR Win Rate use unrounded primary outcomes and VCRs,
respectively. Both metrics average $w_{h,h'}$ equally over opponents and the
included benchmarks, then convert to percentages; ties receive half credit.
Across-model means use the balanced panel. Raw scores use benchmark-specific
scales, while win rates summarize pairwise ordering.

\paragraph{Fit normalization and boundary cases.}
If $J^\star_{m,b}=0$, all ratios in that block are set to zero. All
normalization references in this study are positive in both point estimates
and bootstrap replicates.

Fit is invariant to positive score rescaling within a model--benchmark block
and, with other harnesses fixed, cannot decrease when its own joint score
improves. Changing the candidate panel can change its best-score references.
The geometric cross-model mean is at most the arithmetic mean, with equality
when all model-specific Fits are equal; any zero Fit makes the index zero.

\paragraph{Joint-score decomposition.}
When VCR is zero, the conditional mean within valid completions is undefined,
while $J$ remains zero. Both contributions in
\Cref{app:joint-continuous,fig:joint-score-contributions} divide by all scheduled
attempts rather than clean or non-clean group counts. Conditional means
within either group are undefined for empty groups. Joint and official
scores in \Cref{tab:fit-ablation} also use scheduled attempts as the denominator.

\paragraph{Diagnostic aggregation.}
For the interaction residuals, the same centering within each benchmark gives
residuals whose equal-weight mean equals the aggregate residual in
\Cref{eq:pairing-affinity}. The execution-stage summaries in
\Cref{fig:failure-stage-decomposition} are macro-averages: six benchmarks per
harness in panel (a) and four harnesses per benchmark in panel (b). Harness
summaries weight benchmarks equally.

\paragraph{Fit variants and held-out selection.}
The outcome-only variant replaces $US$ with $S$, retaining normalization and
geometric aggregation; the arithmetic variant replaces the geometric mean
while retaining joint scores and normalization. All benchmarks and models
receive equal weight. For \Cref{tab:fit-sensitivity}, reference scores and
aggregation are recomputed on the retained panel; rankings use unrounded
estimates. In the leave-one-model-out diagnostic, Outcome Win Rate and VCR
Win Rate average model-conditioned rates; outcome-only and joint Fit use
geometric aggregation.

\paragraph{Relative token load.}
Relative load is the geometric mean of within-block normalized Token Loads
across matched model--benchmark blocks.

\Needspace{0.14\textheight}
\section{Cross-Benchmark Comparison Tables}
\label{app:crossbenchmark-comparisons}

\subsection{Outcome Win Rate}
\label{app:outcome-win-rates}

\Cref{tab:outcome-win-rate} reports the exact Outcome Win Rates summarized
in \Cref{sec:results-task-performance}.

\begin{table}[!htbp]
\centering
\small
\setlength{\tabcolsep}{3.2pt}
\caption{Six-benchmark Outcome Win Rates (\%; \Cref{tab:main-results-summary}).
Columns: model-conditioned rates, head-to-head rates pooled across models,
and the three-model mean. DA/OC: DeepAgents/OpenCode; ties receive half credit.
Rates report tie-aware wins across the specified configuration comparisons.}
\label{tab:outcome-win-rate}
\begin{tabular}{lrrr@{\hspace{7pt}}rrrrr}
\toprule
\multirow{2}{*}{Harness}
& \multicolumn{3}{c}{By model}
& \multicolumn{4}{c}{Head-to-head}
& \multirow{2}{*}{\makecell{Overall\\win rate}} \\
\cmidrule(lr){2-4}\cmidrule(lr){5-8}
& \makecell{MiniMax\\M3} & \makecell{Kimi\\K2.7}
& \makecell{Qwen3.6\\35B-A3B}
& vs.\ DA & vs.\ OC & vs.\ Pi & vs.\ Codex & \\
\midrule
DeepAgents & 38.89 & 38.89 & 27.78
           & -- & 22.22 & 33.33 & \underline{50.00} & 35.19 \\
OpenCode   & \textbf{83.33} & \textbf{72.22} & \underline{61.11}
           & \textbf{77.78} & -- & \textbf{66.67} & \textbf{72.22}
           & \textbf{72.22} \\
Pi CLI     & \underline{44.44} & \underline{55.56} & 47.22
           & \underline{66.67} & \textbf{33.33} & -- & 47.22
           & \underline{49.07} \\
Codex CLI  & 33.33 & 33.33 & \textbf{63.89}
           & 50.00 & \underline{27.78} & \underline{52.78} & -- & 43.52 \\
\bottomrule
\end{tabular}
\end{table}

\FloatBarrier
\Needspace{0.14\textheight}

\subsection{Model-Conditioned Valid Completion}
\label{app:model-conditioned-operability}

\Cref{tab:matched-outcome-operability} reports the exact VCR Win Rate and Mean VCR values summarized in \Cref{sec:results-crossbench-operability}.

\begin{table}[!htbp]
\centering
\small
\setlength{\tabcolsep}{6pt}
\renewcommand{\arraystretch}{1.10}
\caption{Six-benchmark model-conditioned VCR Win Rate and Mean VCR
(\%). VCR Win Rate uses tie-aware comparisons; Mean VCR weights benchmarks
equally. Bold/underlined: best/second-best harness within each
column.}
\label{tab:matched-outcome-operability}
\begin{tabular}{@{}
  >{\raggedright\arraybackslash}p{0.15\textwidth}
  *{6}{>{\centering\arraybackslash}p{0.108\textwidth}}
  @{}}
\toprule
\multirow{2}{*}{Harness}
& \multicolumn{2}{c}{MiniMax M3}
& \multicolumn{2}{c}{Kimi K2.7}
& \multicolumn{2}{c}{Qwen3.6-35B-A3B} \\
\cmidrule(lr){2-3}\cmidrule(lr){4-5}\cmidrule(lr){6-7}
& \makecell{VCR\\win rate} & \makecell{Mean\\VCR}
& \makecell{VCR\\win rate} & \makecell{Mean\\VCR}
& \makecell{VCR\\win rate} & \makecell{Mean\\VCR} \\
\midrule
DeepAgents
& 25.00 & 60.60
& 27.78 & 59.69
& 16.67 & 79.31 \\
OpenCode
& \textbf{86.11} & \textbf{91.68}
& \underline{55.56} & \underline{83.09}
& \textbf{77.78} & \textbf{90.70} \\
Pi CLI
& \underline{47.22} & \underline{72.96}
& \textbf{77.78} & \textbf{87.94}
& 50.00 & 82.13 \\
Codex CLI
& 41.67 & 71.88
& 38.89 & 78.26
& \underline{55.56} & \underline{89.48} \\
\bottomrule
\end{tabular}
\end{table}

\FloatBarrier

\subsection{Fixed System References}
\label{app:native-system-reference}

The fixed references provide complete-system baselines, while the crossed
indices compare the factorial model--harness panel.
The GPT-5.5--Codex reference covers all six benchmarks; DeepSeek V4 Pro
0813--DeepSeek Harness uses its standard preset on SWE-bench-45,
Terminal-Bench 2, VitaBench, and LOCA-Bench. Both use the core panels' task
sets, repetitions, external controls, and evaluators, including strict tool
permissions on VitaBench.

GPT-5.5--Codex wins most matched outcome comparisons
across the six benchmarks, but does not lead every benchmark in either outcome
or VCR. DeepSeek V4 Pro 0813--DeepSeek Harness exceeds all 16 core
configurations in Mean-of-4 on SWE-bench-45, Terminal-Bench 2, and VitaBench,
but trails MiniMax--Codex on LOCA-Bench. Complete-system rankings vary by benchmark in both outcome and valid completion. \Cref{tab:native-reference-comparison}
reports the complete comparisons.

The fixed references show that outcome and operational validity remain separate even
for native systems. GPT-5.5--Codex is fully clean on VitaBench without
leading in outcome. On Terminal-Bench 2, it leads every core configuration in
outcome but leaves more than a quarter of attempts non-clean; on ClawMark,
it leads in outcome but is less clean than five core configurations.
DeepSeek V4 Pro 0813--DeepSeek Harness is clean on every SWE attempt, yet 53
remain unresolved. On Terminal-Bench 2, it also leads every core
configuration in outcome, but its VCR trails most of them because
more than a third of its runs terminate abnormally. On VitaBench, it leads
every core configuration in outcome while remaining fully clean. On
LOCA-Bench, both its VCR and Mean-of-4 trail MiniMax--Codex.
The native references exhibit benchmark-specific outcome and valid-completion profiles.

\Cref{tab:native-reference-comparison} compares each fixed reference with
the full set of core configurations on exactly matched task support. For
outcome or VCR $z$, the fixed-reference win rate is
\[
R_{r}^{z}
=
\frac{100}{|\mathcal B_r|}
\sum_{b\in\mathcal B_r}
\frac{1}{|\mathcal C_b|}
\sum_{c\in\mathcal C_b}
\phi(z_{br},z_{bc}),
\]
where $\mathcal C_b$ is the set of evaluated core configurations and ties
receive half credit. Joint Pareto dominance requires the reference to be no
worse in both outcome and VCR and strictly better in at least one. Each
benchmark receives equal weight. These quantities locate each complete system
relative to the evaluated panel.

\begin{table}[!htbp]
\centering
\scriptsize
\setlength{\tabcolsep}{4pt}
\renewcommand{\arraystretch}{1.12}
\caption{Fixed references versus evaluated core configurations on matched tasks.
Outcome: Mean-of-4 (continuous for ClawMark) or AgentSkillOS Task Score.
W/T/L: configuration-level wins/ties/losses; WR: tie-aware reference win rate;
Joint: percentage of configurations Pareto-dominated in both outcome and
VCR. Summaries weight benchmarks equally: six for GPT-5.5--Codex;
SWE-bench-45, Terminal-Bench 2, VitaBench, and LOCA-Bench for DeepSeek Harness.}
\label{tab:native-reference-comparison}
\begin{tabular}{lrrrrrrrr}
\toprule
Benchmark & $|\mathcal C_b|$
& \makecell{Native\\outcome} & \makecell{Outcome\\W/T/L}
& \makecell{Outcome\\WR}
& \makecell{Native\\VCR} & \makecell{VCR\\W/T/L}
& \makecell{VCR\\WR} & Joint \\
\midrule
\multicolumn{9}{l}{\textit{GPT-5.5--Codex}} \\
SWE-bench-45
& 16 & 65.56 & 16/0/0 & 100.00
& 99.44 & 13/3/0 & 90.63 & 100.00 \\
Terminal-Bench 2
& 16 & 60.39 & 16/0/0 & 100.00
& 72.47 & 10/0/6 & 62.50 & 62.50 \\
VitaBench
& 16 & 46.56 & 11/0/5 & 68.75
& 100.00 & 4/12/0 & 62.50 & 68.75 \\
LOCA-Bench
& 16 & 76.00 & 15/0/1 & 93.75
& 96.67 & 8/0/8 & 50.00 & 50.00 \\
ClawMark
& 12 & 74.31 & 12/0/0 & 100.00
& 92.25 & 7/0/5 & 58.33 & 58.33 \\
AgentSkillOS
& 12 & 97.57 & 12/0/0 & 100.00
& 100.00 & 12/0/0 & 100.00 & 100.00 \\
\midrule
Equal-benchmark mean
& 88 & -- & 82/0/6 & 93.75
& -- & 54/15/19 & 70.66 & 73.26 \\
\midrule
\multicolumn{9}{l}{\textit{DeepSeek V4 Pro 0813--DeepSeek Harness}} \\
SWE-bench-45
& 16 & 70.56 & 16/0/0 & 100.00
& 100.00 & 16/0/0 & 100.00 & 100.00 \\
Terminal-Bench 2
& 16 & 46.91 & 16/0/0 & 100.00
& 62.92 & 4/0/12 & 25.00 & 25.00 \\
VitaBench
& 16 & 54.13 & 16/0/0 & 100.00
& 100.00 & 4/12/0 & 62.50 & 100.00 \\
LOCA-Bench
& 16 & 73.33 & 15/0/1 & 93.75
& 92.67 & 7/0/9 & 43.75 & 43.75 \\
\midrule
Equal-benchmark mean
& 64 & -- & 63/0/1 & 98.44
& -- & 31/12/21 & 57.81 & 67.19 \\
\bottomrule
\end{tabular}
\end{table}

On SWE-bench-45, DeepSeek V4 Pro 0813--DeepSeek Harness is five Mean-of-4
points above GPT-5.5--Codex, but the paired task-cluster bootstrap interval
includes zero. On VitaBench, its 7.56-point advantage over GPT-5.5--Codex has a
95\% interval of $[4.19,10.94]$, while both systems have a 100\% VCR.
The outcome difference therefore remains after VCR saturates.
On SWE-bench-45, the two systems also
retain complementary task coverage: DeepSeek Harness uniquely solves four
tasks and GPT-5.5--Codex uniquely solves three.

LOCA-Bench provides a contrasting native-system result. DeepSeek
Harness trails both GPT-5.5--Codex and MiniMax--Codex in Mean-of-4 and VCR on
this benchmark. These native pairings provide fixed references for benchmark-specific outcome and
valid-completion comparisons.

\FloatBarrier

\section{VCR Summaries and Sensitivity}
\label{app:operability-sensitivity}

To complement mean summaries, we report the type-7 25th percentile and minimum
of VCR. We also obtain model-side VCR win rates by comparing models within
benchmark--harness blocks.

The type-7 lower quartile interpolates between adjacent observations in the
fixed evaluation panel. As a discrete sensitivity, we instead take the
nearest-rank empirical quartile---the $\lceil 0.25n\rceil$th ordered observation.
This change preserves both orderings: OpenCode remains the leading harness and
Qwen the leading model (\Cref{tab:nearest-rank-quartiles}). We retain the type-7
statistic as the interpolated lower-quartile summary and report the observed
minimum to represent worst-configuration VCR.

\begin{table}[!htbp]
\centering
\small
\caption{Nearest-rank lower quartiles of VCR (\%). Harness and model
orderings match the type-7 summaries.}
\label{tab:nearest-rank-quartiles}
\begin{tabular}{lrlr}
\toprule
Harness & Lower quartile & Model & Lower quartile \\
\midrule
DeepAgents & 49.16 & MiniMax M3 & 48.60 \\
OpenCode & 81.67 & Kimi K2.7 & 60.00 \\
Pi CLI & 64.33 & Qwen3.6-35B-A3B & 76.00 \\
Codex CLI & 68.82 & & \\
\bottomrule
\end{tabular}
\end{table}

\section{Benchmark-Specific Results}
\label{app:benchmark-results}

\suppressfloats[t]

The following sections retain the complete benchmark-level tables and
benchmark-specific analyses underlying the cross-benchmark findings in the
main text.

\FloatBarrier
\subsection{SWE-bench-45}

\Cref{tab:swebench45-main-results} reports the complete SWE-bench-45
results. The gap between Best-of-4 and Mean-of-4 is substantial in several
configurations. MiniMax M3 with OpenCode, for example, attains the highest Best-of-4 and Mean-of-4 in
its model block. The 19.4-point gap distinguishes success in at least one trial from
average success across all four executions. Harness rankings are model dependent: Codex CLI attains the
highest Mean-of-4 for DeepSeek V4 Flash Preview and Qwen3.6-35B-A3B, and OpenCode for
MiniMax M3 and Kimi K2.7.

\begin{table*}[!htbp]
\centering
\small
\setlength{\tabcolsep}{6pt}
\renewcommand{\arraystretch}{1.10}
\caption{SWE-bench-45 Best-of-4 and Mean-of-4 success rates
(\%; \Cref{app:benchmark-metrics}). Bold/underlined: best/second-best
within each core model, including ties. Fixed references are excluded from
highlighting.}
\label{tab:swebench45-main-results}
\begin{tabular}{llrr}
\toprule
Model & Harness & Best & Mean \\
\midrule
\multirow{4}{*}{\makecell[l]{DeepSeek V4\\Flash Preview}}
 & DeepAgents & 40.00 & 31.11 \\
 & OpenCode   & 42.22 & 32.22 \\
 & Pi CLI     & \textbf{46.67} & \underline{35.56} \\
 & Codex CLI  & \textbf{46.67} & \textbf{36.11} \\
\midrule
\multirow{4}{*}{MiniMax M3}
 & DeepAgents & 55.56 & 31.11 \\
 & OpenCode   & \textbf{82.22} & \textbf{62.78} \\
 & Pi CLI     & \underline{71.11} & \underline{52.22} \\
 & Codex CLI  & 55.56 & 30.56 \\
\midrule
\multirow{4}{*}{Kimi K2.7}
 & DeepAgents & 31.11 & 15.56 \\
 & OpenCode   & \textbf{66.67} & \textbf{48.89} \\
 & Pi CLI     & \underline{51.11} & \underline{42.22} \\
 & Codex CLI  & 28.89 & 10.56 \\
\midrule
\multirow{4}{*}{\makecell[l]{Qwen3.6-\\35B-A3B}}
 & DeepAgents & 31.11 & 15.00 \\
 & OpenCode   & \underline{42.22} & 22.22 \\
 & Pi CLI     & 33.33 & \underline{25.00} \\
 & Codex CLI  & \textbf{44.44} & \textbf{26.67} \\
\midrule
\multicolumn{2}{l}{GPT-5.5 xhigh + Codex CLI} & 86.67 & 65.56 \\
\cmidrule(lr){1-4}
\multicolumn{2}{l}{DeepSeek V4 Pro 0813 + DeepSeek Harness} & 88.89 & 70.56 \\
\bottomrule
\end{tabular}
\end{table*}

\paragraph{SWE-bench-45 operational validity.}
The four-model panel contains 2,880 scheduled SWE executions.
The fixed GPT-5.5--Codex and DeepSeek V4 Pro 0813--DeepSeek Harness references each contribute a
separate 180-run panel on the same 45 tasks.
Here a valid submission is a non-empty patch accepted by the evaluator's
patch-application stage; correctness is excluded from operational validity.
\Cref{tab:swe-operability-matrix} reports the single-benchmark VCR
matrix and its within-SWE model and harness summaries.

\begin{table*}[!htbp]
\centering
\scriptsize
\setlength{\tabcolsep}{4pt}
\caption{SWE-bench-45 VCR (\%). A: four-model matrix and fixed
references. B: core-panel harness/model pairwise win rates, means, type-7
25th percentiles, and minima. These single-benchmark summaries differ from the
balanced three-model, six-benchmark summaries in
\Cref{tab:matched-outcome-operability}.}
\label{tab:swe-operability-matrix}
\textit{Panel A: VCR matrix}\\[2pt]
\begin{tabular}{lrrrrr}
\toprule
Model & DeepAgents & OpenCode & Pi CLI & Codex CLI & \makecell{DeepSeek\\Harness} \\
\midrule
DeepSeek V4 Flash Preview & 99.44 & 99.44 & 94.44 & 99.44 & -- \\
MiniMax M3        & 47.78 & 95.00 & 87.22 & 70.56 & -- \\
Kimi K2.7         & 25.56 & 95.00 & 96.11 & 39.44 & -- \\
Qwen3.6-35B-A3B   & 80.56 & 88.89 & 92.22 & 98.33 & -- \\
\midrule
GPT-5.5 xhigh      & -- & -- & -- & 99.44 & -- \\
DeepSeek V4 Pro 0813    & -- & -- & -- & -- & 100.00 \\
\bottomrule
\end{tabular}
\vspace{4pt}

\textit{Panel B: single-benchmark pairwise and magnitude summaries}\\[2pt]
\begin{tabular}{llrrrr}
\toprule
Comparison & Name & \makecell{Pairwise\\win rate} & Mean & \makecell{25th\\pct.} & Minimum \\
\midrule
\multirow{4}{*}{Harness}
 & DeepAgents & 16.67 & 63.33 & 42.22 & 25.56 \\
 & OpenCode   & 66.67 & 94.58 & 93.47 & 88.89 \\
 & Pi CLI     & 58.33 & 92.50 & 90.97 & 87.22 \\
 & Codex CLI  & 58.33 & 76.94 & 62.78 & 39.44 \\
\midrule
\multirow{4}{*}{Model}
 & DeepSeek V4 Flash Preview & 91.67 & 98.19 & 98.19 & 94.44 \\
 & MiniMax M3        & 29.17 & 75.14 & 64.86 & 47.78 \\
 & Kimi K2.7         & 37.50 & 64.03 & 35.97 & 25.56 \\
 & Qwen3.6-35B-A3B   & 41.67 & 90.00 & 86.81 & 80.56 \\
\bottomrule
\end{tabular}
\end{table*}

DeepSeek combines near-ceiling operational validity across harnesses with a
best Mean-of-4 success rate of 36.11\%. Among the other models,
OpenCode and Pi CLI produce clean
evaluator-ready completions in more than 87\% of runs for every model.
Kimi has the lowest VCR under DeepAgents. Codex CLI shows the opposite
model dependence: its VCR is 58.9 percentage points higher for Qwen
than for Kimi.
Twenty-five runs that ultimately solve the issue are
not counted as valid completions because they terminate abnormally,
illustrating why evaluator handoff and task success are distinct.
Both fixed references combine near-ceiling VCR with substantial
Best-of-4--Mean-of-4 gaps, separating operational validity from repeatable
issue resolution.

\begin{table}[!htbp]
\centering
\small
\caption{SWE-bench-45 additive VCR interaction residual (percentage
points), centered over the four-model SWE panel.}
\label{tab:swe-pairing-affinity}
\begin{tabular}{lrrrr}
\toprule
Model & DeepAgents & OpenCode & Pi CLI & Codex CLI \\
\midrule
DeepSeek V4 Flash Preview & 19.76 & -11.49 & -14.41 & 6.15 \\
MiniMax M3        & -8.85 & 7.12 & 1.42 & 0.31 \\
Kimi K2.7         & -19.97 & 18.23 & 21.42 & -19.69 \\
Qwen3.6-35B-A3B   & 9.06 & -13.85 & -8.44 & 13.23 \\
\bottomrule
\end{tabular}
\end{table}

The trajectory evidence localizes mechanisms consistent with these residuals.
DeepAgents records nearly twice as many \texttt{RuntimeError} failures for
Kimi as for MiniMax on the same scheduled support, while Qwen has none.
Codex CLI terminates normally in all Qwen runs and produces an evaluator-ready
patch in 106 more executions than for Kimi out of the same 180 attempts. These
SWE-specific residuals guide trajectory inspection and depend on the comparison
panel. Their magnitudes attenuate in the matched three-model calculation in
\Cref{fig:crossbench-pairing-affinity}; including DeepSeek changes the centering
baseline and yields the four-model values reported here.

\FloatBarrier
\subsection{Terminal-Bench 2}

\begin{table*}[!htbp]
  \centering
  \small
  \setlength{\tabcolsep}{4.2pt}
  \renewcommand{\arraystretch}{1.12}
  \caption{Terminal-Bench 2 results on all 89 tasks: four models, four harnesses,
  and fixed references. Best/Mean: Best-of-4/Mean-of-4.}
  \label{tab:tb2-4x4-pass-mean}
  \begin{tabular}{lrrrrrrrr}
    \toprule
    \multirow{2}{*}{Model}
      & \multicolumn{2}{c}{DeepAgents}
      & \multicolumn{2}{c}{OpenCode}
      & \multicolumn{2}{c}{PI}
      & \multicolumn{2}{c}{Codex} \\
    \cmidrule(lr){2-3}\cmidrule(lr){4-5}\cmidrule(lr){6-7}\cmidrule(lr){8-9}
      & Best & Mean
      & Best & Mean
      & Best & Mean
      & Best & Mean \\
    \midrule
    Qwen3.6-35BA3B
      & 40.4 & 26.4
      & 40.4 & 26.1
      & \underline{43.8} & \textbf{28.7}
      & \textbf{47.2} & \textbf{28.7} \\
    DeepSeek
      & \textbf{61.8} & \textbf{45.8}
      & 53.9 & 40.4
      & \textbf{61.8} & 44.7
      & 50.6 & 37.6 \\
    MiniMax-M3
      & 46.1 & 21.6
      & \textbf{55.1} & \textbf{39.6}
      & 53.9 & 33.1
      & 51.7 & 34.6 \\
    Kimi-K2.7
      & 44.9 & 30.6
      & 53.9 & 39.0
      & \textbf{59.6} & \textbf{43.8}
      & 48.3 & 31.7 \\
    \bottomrule
  \end{tabular}
  \vspace{5pt}

  \begin{tabular}{lrr}
    \toprule
    Fixed system reference & Best-of-4 (\%) & Mean-of-4 (\%) \\
    \midrule
    GPT-5.5 xhigh--Codex CLI & 76.40 & 60.39 \\
    DeepSeek V4 Pro 0813--DeepSeek Harness & 62.92 & 46.91 \\
    \bottomrule
  \end{tabular}
\end{table*}

Across the full 89 tasks in Terminal-Bench 2, performance varies substantially
across model--harness configurations. Pi CLI has the strongest four-model
macro-average, exceeding runner-up OpenCode by 1.3 percentage points in
Mean-of-4. Its largest model-specific advantage is on Kimi, where it leads
both metrics and exceeds OpenCode by 4.8 points in Mean-of-4. For Qwen,
Codex CLI leads Best-of-4 and ties Pi CLI on Mean-of-4. The
ordering changes for the other models:
DeepAgents leads DeepSeek Mean-of-4 and ties Pi CLI on Best-of-4, whereas OpenCode
leads both MiniMax M3 metrics. The leading harness varies across models.

\Needspace{0.18\textheight}
From the model perspective, DeepSeek V4 Flash Preview remains strongest overall,
while Qwen obtains the lowest averages. Harness choice still affects every
model: for MiniMax, OpenCode exceeds DeepAgents by 18.0 percentage points in
Mean-of-4. The large
Best-of-4--Mean-of-4 gaps distinguish occasional task success from consistent
performance across repeated executions.

The harness ranking is model dependent: Codex CLI leads Qwen Best-of-4 and
ties Pi CLI on Mean-of-4; Pi CLI leads Kimi; DeepAgents leads DeepSeek
Mean-of-4 and ties Pi CLI on Best-of-4; OpenCode leads MiniMax. These reversals
characterize model-dependent performance across full system configurations
that combine prompts, adapters, action-schema formatting, context management,
and runtime settings.

Both fixed references exceed every core configuration in outcome.
GPT-5.5--Codex leads DeepSeek V4 Pro 0813--DeepSeek Harness by 13.5 percentage
points in both Best-of-4 and Mean-of-4. Neither system is fully clean.
More than a third of DeepSeek Harness runs terminate abnormally, while its
final-verifier handoff rate is near ceiling; its valid-completion failures therefore
arise almost entirely from abnormal termination. Seven non-normal runs still receive full
task credit. Terminal-Bench 2 shows that official task scores and execution-validity metrics capture
distinct aspects of system performance.

\FloatBarrier
\subsection{VitaBench}
\Cref{tab:vitabench-strict-results} reports Best-of-4 and Mean-of-4 under strict
tool permissions. Cross-domain interaction is the most difficult
setting under both metrics for every model--harness configuration. In the core
panel, the best Cross configuration falls 20.8 percentage points below the
best single-domain configuration in Mean-of-4; the Best-of-4 gap is 13 points.
Harness effects remain model dependent. By Total Mean-of-4, Codex CLI leads
for DeepSeek, DeepAgents for MiniMax, OpenCode for Kimi, and Pi CLI for Qwen.
By Total Best-of-4, Codex CLI leads for DeepSeek, MiniMax, and Kimi, while Pi
CLI leads for Qwen.

Across the core configurations, the gap between unrounded Total Best-of-4 and
Total Mean-of-4 is roughly 23--32 percentage points. Thus, the configurations
solve a substantially broader set of tasks at least once than they solve
reliably across repeated executions.
The fixed DeepSeek V4 Pro 0813--DeepSeek Harness reference has a 25.1-point
Best-of-4--Mean-of-4 gap, and its Mean-of-4 exceeds every core configuration and
GPT-5.5--Codex. Both references are
excluded from the within-model formatting.

\begin{table*}[!htbp]
\centering
\scriptsize
\setlength{\tabcolsep}{2.2pt}
\renewcommand{\arraystretch}{1.08}
\caption{VitaBench Best-of-4/Mean-of-4 success rates (\%) under strict tool
permissions: 100 tasks and four independent trials (400 attempts) per
environment (\Cref{app:benchmark-metrics}). Cross combines Delivery,
In-store, and OTA; Total averages the four environments equally.
Bold/underlined: best/second-best per core model and metric, including ties.
Fixed references are excluded from highlighting.}
\label{tab:vitabench-strict-results}
\begin{tabular}{llrrrrrrrrrr}
\toprule
\multirow{2}{*}{Model} & \multirow{2}{*}{Harness}
 & \multicolumn{2}{c}{Delivery}
 & \multicolumn{2}{c}{In-store}
 & \multicolumn{2}{c}{OTA}
 & \multicolumn{2}{c}{Cross}
 & \multicolumn{2}{c}{Total} \\
\cmidrule(lr){3-4}\cmidrule(lr){5-6}\cmidrule(lr){7-8}
\cmidrule(lr){9-10}\cmidrule(lr){11-12}
 & & Best & Mean & Best & Mean & Best & Mean
 & Best & Mean & Best & Mean \\
\midrule
MiniMax M3 & DeepAgents & 81.00 & \textbf{59.50} & 83.00 & \underline{58.25} & \underline{76.00} & 47.00 & \underline{69.00} & \underline{39.25} & \underline{77.25} & \textbf{51.00} \\
 & OpenCode & 83.00 & 59.00 & \underline{84.00} & 56.75 & \underline{76.00} & \underline{49.00} & 59.00 & 32.75 & 75.50 & \underline{49.38} \\
 & Pi CLI & \textbf{86.00} & \underline{59.25} & \textbf{87.00} & \textbf{58.50} & \textbf{77.00} & \textbf{49.25} & 59.00 & 30.00 & \underline{77.25} & 49.25 \\
 & Codex CLI & \underline{85.00} & 53.25 & 82.00 & 48.50 & 67.00 & 35.25 & \textbf{76.00} & \textbf{43.25} & \textbf{77.50} & 45.06 \\
\midrule
Kimi K2.7 & DeepAgents & 75.00 & \underline{51.50} & 78.00 & \underline{52.25} & \underline{69.00} & \textbf{47.00} & 42.00 & \underline{21.50} & 66.00 & \underline{43.06} \\
 & OpenCode & \underline{76.00} & \textbf{54.50} & 83.00 & \textbf{56.00} & 62.00 & 42.25 & \textbf{53.00} & \textbf{24.75} & 68.50 & \textbf{44.38} \\
 & Pi CLI & 75.00 & 48.75 & \underline{84.00} & \underline{52.25} & 67.00 & 40.75 & \underline{49.00} & 19.25 & \underline{68.75} & 40.25 \\
 & Codex CLI & \textbf{79.00} & 50.75 & \textbf{89.00} & 50.50 & \textbf{72.00} & \underline{44.00} & 48.00 & 20.50 & \textbf{72.00} & 41.44 \\
\midrule
Qwen3.6-35B-A3B & DeepAgents & \underline{74.00} & 39.75 & \textbf{75.00} & 35.25 & \textbf{60.00} & 26.00 & 28.00 & 9.75 & \underline{59.25} & 27.69 \\
 & OpenCode & 69.00 & \underline{41.25} & 72.00 & \underline{37.25} & \underline{55.00} & \textbf{29.50} & \textbf{39.00} & \textbf{15.50} & 58.75 & \underline{30.88} \\
 & Pi CLI & \textbf{76.00} & \textbf{44.75} & 73.00 & \textbf{38.00} & 53.00 & \underline{27.75} & \underline{36.00} & \underline{14.25} & \textbf{59.50} & \textbf{31.19} \\
 & Codex CLI & 68.00 & 35.50 & \underline{74.00} & 31.25 & 42.00 & 17.25 & 23.00 & 7.50 & 51.75 & 22.88 \\
\midrule
DeepSeek V4 Flash Preview & DeepAgents & 78.00 & 46.75 & 80.00 & 48.25 & \underline{73.00} & 39.25 & 45.00 & 17.75 & 69.00 & 38.00 \\
 & OpenCode & 83.00 & 58.75 & \underline{86.00} & 52.00 & 65.00 & 42.00 & 52.00 & \underline{28.75} & 71.50 & 45.38 \\
 & Pi CLI & \underline{85.00} & \underline{60.50} & \textbf{87.00} & \textbf{53.75} & 70.00 & \textbf{46.75} & \underline{61.00} & \textbf{34.75} & \underline{75.75} & \underline{48.94} \\
 & Codex CLI & \textbf{86.00} & \textbf{64.00} & \underline{86.00} & \underline{52.75} & \textbf{74.00} & \underline{44.75} & \textbf{62.00} & \textbf{34.75} & \textbf{77.00} & \textbf{49.06} \\
\midrule
\multicolumn{2}{l}{GPT-5.5 xhigh + Codex CLI}
 & 81.00 & 58.75 & 80.00 & 51.50 & 71.00 & 48.75
 & 44.00 & 27.25 & 69.00 & 46.56 \\
\cmidrule(lr){1-12}
\multicolumn{2}{l}{DeepSeek V4 Pro 0813 + DeepSeek Harness}
 & 87.00 & 61.00 & 90.00 & 62.75 & 73.00 & 52.00
 & 67.00 & 40.75 & 79.25 & 54.13 \\
\bottomrule
\end{tabular}
\end{table*}

\paragraph{VitaBench operational validity.}
The four-model strict panel contains 25,600 scheduled attempts.
The DeepSeek Harness reference contributes a separate 1,600-run panel on the
same task--trial support and reaches a 100\% VCR.
\Cref{tab:vitabench-operability} reports the complete single-benchmark
process profile.

\begin{table*}[!htbp]
\centering
\scriptsize
\setlength{\tabcolsep}{3.2pt}
\caption{VitaBench process metrics under strict permissions. A: VCR (\%)
and model/harness summaries. B: four-model interaction residuals (pp).
C: harness-level normal termination, valid handoff, and VCR
(\%). D: fixed references.}
\label{tab:vitabench-operability}
\textit{Panel A: VCRs with model and harness summaries}\\[2pt]
\begin{tabular}{lrrrrrrrr}
\toprule
Model & DeepAgents & OpenCode & Pi CLI & Codex CLI
& Mean & \makecell{25th\\pct.} & Minimum & \makecell{Model-side\\win rate} \\
\midrule
DeepSeek V4 Flash Preview & 100.00 & 99.94 & 100.00 & 100.00 & 99.98 & 99.98 & 99.94 & 50.00 \\
MiniMax M3        & 100.00 & 100.00 & 100.00 & 100.00 & 100.00 & 100.00 & 100.00 & 66.67 \\
Kimi K2.7         & 100.00 & 100.00 & 100.00 & 99.94 & 99.98 & 99.98 & 99.94 & 50.00 \\
Qwen3.6-35B-A3B   & 98.25 & 100.00 & 100.00 & 99.94 & 99.55 & 99.52 & 98.25 & 33.33 \\
\midrule
Harness mean       & 99.56 & 99.98 & 100.00 & 99.97 & -- & -- & -- & -- \\
Harness 25th pct.  & 99.56 & 99.98 & 100.00 & 99.94 & -- & -- & -- & -- \\
Harness minimum    & 98.25 & 99.94 & 100.00 & 99.94 & -- & -- & -- & -- \\
VCR win rate      & 45.83 & 50.00 & 66.67 & 37.50 & -- & -- & -- & -- \\
\bottomrule
\end{tabular}
\vspace{4pt}

\textit{Panel B: additive interaction residual}\\[2pt]
\begin{tabular}{lrrrr}
\toprule
Model & DeepAgents & OpenCode & Pi CLI & Codex CLI \\
\midrule
DeepSeek V4 Flash Preview & +0.33 & -0.15 & -0.11 & -0.07 \\
MiniMax M3        & +0.32 & -0.11 & -0.12 & -0.09 \\
Kimi K2.7         & +0.33 & -0.09 & -0.11 & -0.14 \\
Qwen3.6-35B-A3B   & -0.98 & +0.35 & +0.33 & +0.30 \\
\bottomrule
\end{tabular}
\vspace{4pt}

\textit{Panel C: harness-level termination and handoff components}\\[2pt]
\begin{tabular}{lrrr}
\toprule
Harness & Normal exit & Valid handoff & VCR \\
\midrule
DeepAgents & 99.56 & 100.00 & 99.56 \\
OpenCode   & 99.98 & 100.00 & 99.98 \\
Pi CLI     & 100.00 & 100.00 & 100.00 \\
Codex CLI  & 99.97 & 100.00 & 99.97 \\
\bottomrule
\end{tabular}
\vspace{4pt}

\textit{Panel D: fixed system references}\\[2pt]
\begin{tabular}{lrrr}
\toprule
Configuration & Normal exit & Valid handoff & VCR \\
\midrule
GPT-5.5--Codex & 100.00 & 100.00 & 100.00 \\
\makecell[l]{DeepSeek V4 Pro 0813--\\DeepSeek Harness}
               & 100.00 & 100.00 & 100.00 \\
\bottomrule
\end{tabular}
\end{table*}

VitaBench VCR is near ceiling: every core configuration has a VCR of at
least 98\%, and the within-benchmark additive residuals lie within one
percentage point of zero. Total Mean-of-4 spans 28.1 percentage points across core configurations, and the leading
harness varies by model. VitaBench therefore isolates a regime in which
model--harness outcome differences persist even when almost every run executes
and hands off successfully. Both fixed references achieve 100\% VCR, while DeepSeek Harness leads GPT-5.5--Codex by
7.6 points in Mean-of-4, providing a further example of outcome differences at the same
operational-validity level.

Resource use also shows a model-conditioned difference. For DeepSeek, Pi CLI
uses less than half as many tokens per attempt as Codex CLI, while trailing
it by less than 0.2 percentage points in Mean-of-4.
These fully covered cells use the same accounting rule as the cross-benchmark resource analysis in
\Cref{sec:results-token-efficiency}.

\FloatBarrier
\FloatBarrier
\subsection{LOCA-Bench}
\paragraph{Main results.}

\Cref{tab:loca-main} reports the primary outcomes. Codex CLI is the best
harness within every model block. Its advantage is especially large for
MiniMax M3: Codex CLI exceeds runner-up DeepAgents by 63 percentage points
in Mean-of-4. For each remaining model, its lead over the next-best harness
is roughly 15--20 points.
MiniMax--Codex exceeds GPT-5.5--Codex by 11 Mean-of-4 points and exceeds both
fixed references in Best-of-4.

\begin{table*}[!htbp]
  \centering
  \small
  \setlength{\tabcolsep}{4.2pt}
  \renewcommand{\arraystretch}{1.12}
  \caption{LOCA-Bench 256K Best-of-4/Mean-of-4 success rates (\%) for the four-model,
four-harness panel and fixed references. Bold marks the best harness per
core model and metric; references are excluded.}
  \label{tab:loca-main}
  \begin{tabular}{lrrrrrrrr}
    \toprule
    \multirow{2}{*}{Model}
      & \multicolumn{2}{c}{DeepAgents}
      & \multicolumn{2}{c}{OpenCode}
      & \multicolumn{2}{c}{Pi CLI}
      & \multicolumn{2}{c}{Codex CLI} \\
    \cmidrule(lr){2-3}\cmidrule(lr){4-5}\cmidrule(lr){6-7}\cmidrule(lr){8-9}
      & Best & Mean
      & Best & Mean
      & Best & Mean
      & Best & Mean \\
    \midrule
    DeepSeek V4 Flash Preview
      & 40.00 & 20.67
      & 53.33 & 31.00
      & 42.67 & 29.00
      & \textbf{76.00} & \textbf{50.67} \\
    Qwen 3.6--35B--A3B
      & 14.67 & 8.67
      & 34.67 & 21.00
      & 28.00 & 15.00
      & \textbf{70.67} & \textbf{36.33} \\
    MiniMax M3
      & 33.33 & 24.00
      & 32.00 & 21.00
      & 21.33 & 12.67
      & \textbf{93.33} & \textbf{87.00} \\
    Kimi K2.7
      & 17.33 & 8.33
      & 42.67 & 27.67
      & 50.67 & 34.33
      & \textbf{86.67} & \textbf{54.00} \\
    \bottomrule
  \end{tabular}
  \vspace{5pt}

  \begin{tabular}{lrr}
    \toprule
    Fixed system reference & Best-of-4 (\%) & Mean-of-4 (\%) \\
    \midrule
    GPT-5.5 xhigh--Codex CLI & 90.67 & 76.00 \\
    DeepSeek V4 Pro 0813--DeepSeek Harness & 90.67 & 73.33 \\
    \bottomrule
  \end{tabular}
\end{table*}

\paragraph{Operational validity.}

The four-model panel contains $4\times4\times75\times4=4{,}800$ executions,
and each fixed system reference contributes 300 runs on the same 75 tasks. The
normal-exit condition requires \texttt{status=success}. The
valid-handoff condition requires a finite final-state evaluator result,
including a result recorded before a subsequently abnormal backend exit. A run
is clean only when both conditions hold. \Cref{tab:loca-operability-matrix} reports the resulting four-model panel.

\begin{table*}[!htbp]
\centering
\scriptsize
\setlength{\tabcolsep}{4pt}
\caption{LOCA-Bench VCR (\%). A: four-model matrix; B: fixed references;
C: core-panel harness/model pairwise win rates, means, type-7 25th percentiles,
and minima.}
\label{tab:loca-operability-matrix}
\textit{Panel A: VCR matrix}\\[2pt]
\begin{tabular}{lrrrr}
\toprule
Model & DeepAgents & OpenCode & Pi CLI & Codex CLI \\
\midrule
DeepSeek V4 Flash Preview & 95.33 & 99.67 & 98.00 & 98.67 \\
Qwen3.6-35B-A3B   & 76.00 & 99.67 & 80.33 & 99.33 \\
MiniMax M3        & 32.33 & 97.33 & 39.67 & 98.00 \\
Kimi K2.7         & 16.33 & 60.00 & 48.00 & 98.00 \\
\bottomrule
\end{tabular}
\vspace{4pt}

\textit{Panel B: fixed system references}\\[2pt]
\begin{tabular}{lrrr}
\toprule
System & Normal exit & Valid output & VCR \\
\midrule
GPT-5.5 xhigh--Codex CLI & 96.67 & 100.00 & 96.67 \\
DeepSeek V4 Pro 0813--DeepSeek Harness & 92.67 & 100.00 & 92.67 \\
\bottomrule
\end{tabular}
\vspace{4pt}

\textit{Panel C: single-benchmark pairwise and magnitude summaries}\\[2pt]
\begin{tabular}{llrrrr}
\toprule
Comparison & Name & \makecell{Pairwise\\win rate} & Mean & \makecell{25th\\pct.} & Minimum \\
\midrule
\multirow{4}{*}{Harness}
 & DeepAgents & 0.00 & 55.00 & 28.33 & 16.33 \\
 & OpenCode   & 83.33 & 89.17 & 88.00 & 60.00 \\
 & Pi CLI     & 33.33 & 66.50 & 45.92 & 39.67 \\
 & Codex CLI  & 83.33 & 98.50 & 98.00 & 98.00 \\
\midrule
\multirow{4}{*}{Model}
 & DeepSeek V4 Flash Preview & 87.50 & 97.92 & 97.33 & 95.33 \\
 & Qwen3.6-35B-A3B   & 79.17 & 88.83 & 79.25 & 76.00 \\
 & MiniMax M3        & 20.83 & 66.83 & 37.83 & 32.33 \\
 & Kimi K2.7         & 12.50 & 55.58 & 40.08 & 16.33 \\
\bottomrule
\end{tabular}
\end{table*}

The interaction is substantially larger than a harness-wide average suggests.
For the same DeepAgents interface, DeepSeek's VCR exceeds Kimi's by
79 percentage points. For MiniMax, Codex CLI exceeds DeepAgents by 65.7
points. DeepAgents' valid-state rate equals its VCR in all four model
configurations, while valid-state rates for the other harnesses are at or
near ceiling. Normal-execution failures account for most LOCA VCR losses; final-state evaluation after
normal stops contributes comparatively fewer losses. The same point is visible in
the four-model double-centered residuals in
\Cref{tab:loca-operability-interaction}.
Both native references have lower VCRs than several core configurations, and
their non-normal runs retain evaluator records with zero task scores.

\begin{table}[!htbp]
\centering
\small
\caption{LOCA-Bench additive VCR interaction residual (percentage
points), centered over the four-model panel.}
\label{tab:loca-operability-interaction}
\begin{tabular}{lrrrr}
\toprule
Model & DeepAgents & OpenCode & Pi CLI & Codex CLI \\
\midrule
DeepSeek V4 Flash Preview & +19.71 & -10.13 & +10.88 & -20.46 \\
Qwen3.6-35B-A3B   & +9.46 & -1.04 & +2.29 & -10.71 \\
MiniMax M3        & -12.21 & +18.63 & -16.38 & +9.96 \\
Kimi K2.7         & -16.96 & -7.46 & +3.21 & +21.21 \\
\bottomrule
\end{tabular}
\end{table}

\paragraph{Harness and model effects.}

The marginal means in \Cref{tab:loca-marginal} show that harness choice
has a larger effect than model choice in this setting. The spread between
harness marginal means is 2.6 times the spread between model marginal means
for Mean-of-4, and 3.5 times for Best-of-4. A sum-of-squares decomposition
likewise assigns most aggregate configuration variation to the harness main
effect, followed by the interaction and then the model main effect
(\Cref{tab:loca-marginal}, lower panel). These shares describe the distribution of
variation among the aggregated configuration means.

\begin{table}[!htbp]
\centering
\caption{LOCA-Bench marginal outcomes and descriptive shares of configuration
variation (\%). Marginal means average over four models or four harnesses.}
\label{tab:loca-marginal}
\small
\textit{Marginal outcomes}\\[2pt]
\begin{tabular}{llrr}
\toprule
Grouping & Level & Best-of-4 (\%) & Mean-of-4 (\%) \\
\midrule
\multirow{4}{*}{Harness}
 & DeepAgents & 26.33 & 15.42 \\
 & OpenCode & 40.67 & 25.17 \\
 & Pi CLI & 35.67 & 22.75 \\
 & Codex CLI & \textbf{81.67} & \textbf{57.00} \\
\midrule
\multirow{4}{*}{Model}
 & DeepSeek V4 Flash Preview & \textbf{53.00} & 32.83 \\
 & Qwen 3.6--35B--A3B & 37.00 & 20.25 \\
 & MiniMax M3 & 45.00 & \textbf{36.17} \\
 & Kimi K2.7 & 49.33 & 31.08 \\
\bottomrule
\end{tabular}
\vspace{4pt}

\textit{Shares of configuration variation}\\[2pt]
\begin{tabular}{lrrr}
\toprule
Outcome & Harness & Model & Interaction \\
\midrule
Mean-of-4 & 67.2 & 9.4 & 23.4 \\
Best-of-4 & 81.9 & 6.5 & 11.6 \\
\bottomrule
\end{tabular}
\end{table}

The model--harness interaction produces ranking reversals. MiniMax is best
under DeepAgents and Codex CLI, DeepSeek is best under OpenCode, and Kimi is
best under Pi CLI. Model rankings are therefore conditional on the execution framework.

\paragraph{Token efficiency and runtime evidence.}

For DeepSeek, Codex CLI uses about 86\% fewer tokens per attempt than Pi CLI
while increasing Mean-of-4 by 21.7 percentage points. The higher success rate
also widens its resource advantage: tokens per success are about 92\% lower.

Trajectory status makes the system-level interpretation more concrete.
We show a runtime-error comparison on Kimi in \Cref{tab:loca-runtime}. Codex has
114 fewer runtime errors than runner-up OpenCode across the same 300 attempts,
and no timeout-like failures. Some non-Codex runs record an evaluator-correct environment state alongside an error
status, separating final-state correctness from process termination and result
write-back. DeepAgents additionally exhibits recursion-limit and partial-state
recovery events; OpenCode errors in this slice are uniformly timeout-like;
Pi CLI shows timeout-like exits and broken pipes. These observations
characterize LOCA-Bench performance through context organization, tool
protocol, and runtime reliability.

\begin{table}[!htbp]
\centering
\caption{Runtime-error evidence for Kimi K2.7 (300 attempts per harness).}
\label{tab:loca-runtime}
\small
\begin{tabular}{lrrp{0.43\linewidth}}
\toprule
Harness & Runtime errors & Timeout-like failures & Main observed pattern \\
\midrule
Codex CLI & 6 & 0 & Most runs terminate normally. \\
\midrule
Pi CLI & 156 & 154 & Timeouts, and \texttt{BrokenPipeError}; some error runs still have a correct final state. \\
\midrule
OpenCode & 120 & 120 & Timeout termination during long tasks. \\
\midrule
DeepAgents & 251 & 235 & Long timeouts, graph recursion limits, forced completion, and partial-state recovery. \\
\bottomrule
\end{tabular}
\end{table}

Under the 256K, multi-tool, exact-state regime, variation among harness
marginal means exceeds variation among model marginal means. The trajectory
analysis characterizes differences in context selection, tool-result
management, and stopping behavior across the evaluated systems.

\FloatBarrier
\subsection{ClawMark}

\Cref{tab:clawmark-main-results} reports ClawMark results for three models
across four harnesses and the fixed native reference. OpenCode achieves the strongest aggregate core-panel results, with all three models
obtaining high scores. M3 shows the most balanced high-score
profile across harnesses, remaining competitive under OpenCode, DeepAgents,
Pi, and Codex. K2 exhibits the sharpest harness dependence: it achieves the highest Mean-of-4 point
estimate under OpenCode and remains competitive under DeepAgents, with Mean-of-4 scores
of 11.93 under Pi and 11.96 under Codex. Qwen shows broad compatibility across four harnesses, with peak scores below those of
the strongest M3 and K2 configurations. The GPT-5.5--Codex reference exceeds the strongest core
configuration, Kimi--OpenCode, by 4.0 Mean-of-4 score points and also leads in
Best-of-4. These results
show that harness choice directly affects how model capabilities are expressed
during task execution, motivating interpretation at the model--harness level.

\Cref{sec:clawmark-action-realization} examines K2's outcome--validity
discrepancy using strict response-level annotations with fully observable
cross-harness denominators and task-level uncertainty.

The harnesses also differ in how they sustain long-horizon execution. OpenCode generally reaches high scores through relatively compact trajectories, translating model decisions into executable actions and state updates with fewer execution steps.
DeepAgents generally produces longer trajectories with substantially more recorded execution steps. This extended execution increases exposure to turn-level timeouts and external service or transport failures. Some interrupted trajectories still receive substantial rubric credit, indicating that meaningful progress was made before termination. Pi and Codex remain effective for Qwen and M3, with performance depending more strongly on
the model's consistency in following the expected structured-action protocol.

\begin{table*}[!htbp]
\centering
\small
\caption{ClawMark Best-of-4/Mean-of-4 results (\%). Score uses continuous evaluator
scores; Full Success requires a run score of 1. Bold/underlined:
best/second-best per core model and metric, including ties. The fixed native
reference is excluded from highlighting.}
\label{tab:clawmark-main-results}
\begin{tabular}{llrrrr}
\toprule
Model & Harness & \multicolumn{2}{c}{Score} & \multicolumn{2}{c}{Full Success} \\
\cmidrule(lr){3-4}\cmidrule(lr){5-6}
 & & Best & Mean & Best & Mean \\
\midrule
MiniMax M3
 & DeepAgents & 75.06 & 54.78 & 21.00 & 11.00 \\
 & OpenCode   & \textbf{82.39} & \textbf{69.62} & \textbf{28.00} & \underline{12.75} \\
 & Pi CLI     & \underline{78.34} & \underline{62.58} & \textbf{28.00} & \textbf{14.50} \\
 & Codex CLI  & 77.25 & 58.49 & \underline{22.00} & 11.75 \\
\midrule
Kimi K2.7
 & DeepAgents & \underline{72.56} & \underline{58.65} & \underline{20.00} & \underline{8.25} \\
 & OpenCode   & \textbf{81.83} & \textbf{70.36} & \textbf{22.00} & \textbf{10.25} \\
 & Pi CLI     & 12.08 & 11.93 & 0.00 & 0.00 \\
 & Codex CLI  & 12.20 & 11.96 & 0.00 & 0.00 \\
\midrule
Qwen3.6-35B-A3B
 & DeepAgents & 68.58 & 52.71 & 10.00 & 4.50 \\
 & OpenCode   & \underline{76.25} & \textbf{63.62} & \underline{11.00} & \underline{4.75} \\
 & Pi CLI     & \textbf{76.91} & 52.61 & \textbf{20.00} & \textbf{6.50} \\
 & Codex CLI  & 75.19 & \underline{61.21} & 10.00 & \underline{4.75} \\
\midrule
\multicolumn{2}{l}{GPT-5.5 xhigh + Codex CLI}
 & 82.54 & 74.31 & 28.00 & 16.25 \\
\bottomrule
\end{tabular}
\end{table*}

\FloatBarrier

\paragraph{ClawMark operational validity.}
The core panel contains 4,800 scheduled attempts, and the fixed native
reference contributes another 400 attempts on the same 100 tasks. The
valid-handoff rate is 100 in every configuration, so VCR is determined by
normal termination.
\Cref{tab:clawmark-operability} reports the resulting matrix and
single-benchmark interaction residuals.

\begin{table*}[!htbp]
\centering
\scriptsize
\setlength{\tabcolsep}{3.5pt}
\caption{ClawMark process metrics. A: core-panel VCR (\%) with
model/harness summaries and a separate fixed reference. B: core-panel
additive interaction residuals (pp).}
\label{tab:clawmark-operability}
\textit{Panel A: VCRs with model and harness summaries}\\[2pt]
\begin{tabular}{lrrrrrrrr}
\toprule
Model & DeepAgents & OpenCode & Pi CLI & Codex CLI
& Mean & \makecell{25th\\pct.} & Minimum & \makecell{Model-side\\win rate} \\
\midrule
MiniMax M3       & 53.50 & 90.00 & 72.25 & 78.00 & 73.44 & 67.56 & 53.50 & 25.00 \\
Kimi K2.7        & 53.25 & 88.00 & 99.25 & 100.00 & 85.13 & 79.31 & 53.25 & 50.00 \\
Qwen3.6-35B-A3B  & 83.50 & 96.75 & 96.75 & 94.00 & 92.75 & 91.38 & 83.50 & 75.00 \\
\midrule
GPT-5.5 xhigh (native) & -- & -- & -- & 92.25 & -- & -- & -- & -- \\
\midrule
Harness mean       & 63.42 & 91.58 & 89.42 & 90.67 & -- & -- & -- & -- \\
Harness 25th pct.  & 53.38 & 89.00 & 84.50 & 86.00 & -- & -- & -- & -- \\
Harness minimum    & 53.25 & 88.00 & 72.25 & 78.00 & -- & -- & -- & -- \\
VCR win rate      & 0.00 & 72.22 & 61.11 & 66.67 & -- & -- & -- & -- \\
\bottomrule
\end{tabular}
\vspace{4pt}

\textit{Panel B: additive interaction residual}\\[2pt]
\begin{tabular}{lrrrr}
\toprule
Model & DeepAgents & OpenCode & Pi CLI & Codex CLI \\
\midrule
MiniMax M3       & +0.42 & +8.75 & -6.83 & -2.33 \\
Kimi K2.7        & -11.52 & -4.94 & +8.48 & +7.98 \\
Qwen3.6-35B-A3B  & +11.10 & -3.81 & -1.65 & -5.65 \\
\bottomrule
\end{tabular}
\end{table*}

\FloatBarrier

For Kimi, Pi CLI and Codex CLI combine near-ceiling VCR with low Mean-of-4
scores, separating valid completion from task-solving performance.
For MiniMax, OpenCode leads VCR comparisons and exceeds its
additive baseline, whereas for Kimi the positive residuals move to Pi CLI and
Codex CLI. Qwen has the highest model-side VCR win rate, but its largest
positive residual is with DeepAgents. These VCR patterns vary with both the
harness and the model even within one long-horizon workplace benchmark.

The native reference leads Score Mean-of-4 while ranking below five core
configurations in VCR. All 31 non-normal runs retain evaluator scores,
including eight with full success, showing that outcome and termination
metrics capture distinct aspects of execution.

Among configurations that sustain task execution, OpenCode combines the
strongest aggregate outcomes with comparatively compact trajectories. For
MiniMax M3, it uses about 44\% fewer tokens per attempt than DeepAgents.
Kimi K2.7 under Pi CLI and Codex CLI combines low Token Load with low task
scores, illustrating the value of interpreting resource use together with
outcomes.

\FloatBarrier
\subsection{AgentSkillOS}
\Cref{tab:agentskillos-main-results} reports the official evaluator score
for each configuration. The larger skill pool changes the ordering rather than
shifting all configurations uniformly. OpenCode leads
MiniMax and Qwen under the primary 200-skill condition, whereas Pi CLI leads
both models under the 900-skill stress condition. Pi CLI is
strongest for Kimi in both conditions. Skill-pool scale therefore interacts
with both the model and the harness.

\begin{table*}[!htbp]
\centering
\small
\caption{AgentSkillOS mean Task Score over four runs: 200 skills (primary),
900 (matched stress). Bold/underlined: best/second-best per core
model--condition block; fixed reference excluded.}
\label{tab:agentskillos-main-results}
\begin{tabular}{llrr}
\toprule
Model & Harness & \makecell{200 skills\\(primary)} & \makecell{900 skills\\(stress)} \\
\midrule
MiniMax M3 & DeepAgents & 80.07 & \underline{83.02} \\
 & OpenCode & \textbf{93.49} & 64.08 \\
 & Pi CLI & \underline{85.39} & \textbf{89.98} \\
 & Codex CLI & 36.44 & 36.42 \\
\midrule
Kimi K2.7 & DeepAgents & \underline{89.92} & \underline{84.34} \\
 & OpenCode & 84.94 & 62.87 \\
 & Pi CLI & \textbf{95.91} & \textbf{90.87} \\
 & Codex CLI & 55.66 & 66.57 \\
\midrule
Qwen3.6-35B-A3B & DeepAgents & \underline{74.01} & 61.77 \\
 & OpenCode & \textbf{76.12} & 50.15 \\
 & Pi CLI & 56.23 & \textbf{75.98} \\
 & Codex CLI & 63.06 & \underline{71.42} \\
\midrule
\multicolumn{2}{l}{GPT-5.5 xhigh + Codex CLI} & 97.57 & 89.68 \\
\bottomrule
\end{tabular}
\end{table*}

\paragraph{AgentSkillOS operational validity.}
Each skill-pool condition contributes 1,440 scheduled attempts on the balanced
three-model panel. The primary cross-benchmark analysis uses
\texttt{skill\_200}; \texttt{skill\_900} is a stress condition that changes the
visible skill-selection burden while keeping tasks, models, and harnesses
matched. \Cref{tab:agentskillos-operability}
reports the complete process summary for both conditions.

\Cref{tab:agentskillos-pairing-affinity} shows that the stress condition
changes valid completion differently across harnesses. Increasing the visible
skill pool from
200 to 900 reduces OpenCode's model-averaged VCR by 26.7 percentage
points, but increases Pi CLI's by 6.4 points. Their VCR win rates move
in the same opposing directions. At the pair level, Codex CLI operates less
cleanly than the additive baseline predicts for MiniMax, but more cleanly for
Qwen, under both conditions.
The effect of skill-pool size depends jointly on the model and the harness
used for skill selection and execution.

\begin{table*}[!htbp]
\centering
\scriptsize
\setlength{\tabcolsep}{3.2pt}
\caption{AgentSkillOS process metrics (\%). A/B: VCR and
model/harness summaries (pairwise win rates, means, type-7 25th percentiles,
and minima). C: normal-exit and valid-output rates. Native references remain
separate from balanced-panel summaries.}
\label{tab:agentskillos-operability}
\textit{Panel A: \texttt{skill\_200} (primary)}\\[2pt]
\begin{tabular}{lrrrrrrrr}
\toprule
Model & DeepAgents & OpenCode & Pi CLI & Codex CLI
& Mean & \makecell{25th\\pct.} & Minimum & \makecell{Model-side\\win rate} \\
\midrule
MiniMax M3       & 80.83 & 96.67 & 90.00 & 36.67 & 76.04 & 69.79 & 36.67 & 37.50 \\
Kimi K2.7        & 92.50 & 83.33 & 99.17 & 63.33 & 84.58 & 78.33 & 63.33 & 75.00 \\
Qwen3.6-35B-A3B  & 82.50 & 81.67 & 59.17 & 71.67 & 73.75 & 68.54 & 59.17 & 37.50 \\
\midrule
Harness mean       & 85.28 & 87.22 & 82.78 & 57.22 & -- & -- & -- & -- \\
Harness 25th pct.  & 81.67 & 82.50 & 74.58 & 50.00 & -- & -- & -- & -- \\
Harness minimum    & 80.83 & 81.67 & 59.17 & 36.67 & -- & -- & -- & -- \\
VCR win rate      & 66.67 & 66.67 & 55.56 & 11.11 & -- & -- & -- & -- \\
\bottomrule
\end{tabular}
\vspace{4pt}

\textit{Panel B: \texttt{skill\_900} (stress)}\\[2pt]
\begin{tabular}{lrrrrrrrr}
\toprule
Model & DeepAgents & OpenCode & Pi CLI & Codex CLI
& Mean & \makecell{25th\\pct.} & Minimum & \makecell{Model-side\\win rate} \\
\midrule
MiniMax M3       & 86.67 & 64.17 & 92.50 & 41.67 & 71.25 & 58.54 & 41.67 & 50.00 \\
Kimi K2.7        & 85.83 & 65.83 & 94.17 & 70.83 & 79.17 & 69.58 & 65.83 & 75.00 \\
Qwen3.6-35B-A3B  & 71.67 & 51.67 & 80.83 & 79.17 & 70.83 & 66.67 & 51.67 & 25.00 \\
\midrule
Harness mean       & 81.39 & 60.56 & 89.17 & 63.89 & -- & -- & -- & -- \\
Harness 25th pct.  & 78.75 & 57.92 & 86.67 & 56.25 & -- & -- & -- & -- \\
Harness minimum    & 71.67 & 51.67 & 80.83 & 41.67 & -- & -- & -- & -- \\
VCR win rate      & 55.56 & 11.11 & 100.00 & 33.33 & -- & -- & -- & -- \\
\bottomrule
\end{tabular}
\vspace{4pt}

\textit{Panel C: harness-level termination and handoff components}\\[2pt]
\begin{tabular}{llrrr}
\toprule
Condition & Harness & Normal exit & Valid output & VCR \\
\midrule
\multirow{4}{*}{\texttt{skill\_200}}
 & DeepAgents & 93.06 & 85.28 & 85.28 \\
 & OpenCode   & 95.83 & 89.17 & 87.22 \\
 & Pi CLI     & 95.83 & 83.89 & 82.78 \\
 & Codex CLI  & 89.17 & 57.22 & 57.22 \\
\midrule
\multirow{4}{*}{\texttt{skill\_900}}
 & DeepAgents & 95.56 & 81.39 & 81.39 \\
 & OpenCode   & 91.11 & 63.33 & 60.56 \\
 & Pi CLI     & 98.61 & 89.44 & 89.17 \\
 & Codex CLI  & 93.89 & 64.44 & 63.89 \\
\midrule
\texttt{skill\_200} & GPT-5.5--Codex & 100.00 & 100.00 & 100.00 \\
\texttt{skill\_900} & GPT-5.5--Codex & 100.00 & 96.67 & 96.67 \\
\bottomrule
\end{tabular}
\end{table*}

\begin{table*}[!htbp]
\centering
\small
\setlength{\tabcolsep}{7pt}
\caption{AgentSkillOS additive VCR interaction residual (percentage
points). Residuals are centered separately within each skill-pool condition.}
\label{tab:agentskillos-pairing-affinity}
\begin{tabular}{llrrrr}
\toprule
Condition & Model & DeepAgents & OpenCode & Pi CLI & Codex CLI \\
\midrule
\multirow{3}{*}{\texttt{skill\_200}}
 & MiniMax M3      & -2.36 & +11.53 & +9.31 & -18.47 \\
 & Kimi K2.7       & +0.76 & -10.35 & +9.93 & -0.35 \\
 & Qwen3.6-35B-A3B & +1.60 & -1.18 & -19.24 & +18.82 \\
\midrule
\multirow{3}{*}{\texttt{skill\_900}}
 & MiniMax M3      & +7.78 & +6.11 & +5.83 & -19.72 \\
 & Kimi K2.7       & -0.97 & -0.14 & -0.42 & +1.53 \\
 & Qwen3.6-35B-A3B & -6.81 & -5.97 & -5.42 & +18.19 \\
\bottomrule
\end{tabular}
\end{table*}

\FloatBarrier
\subsection{Complete Resource Results}
\label{app:detailed-resource-accounting}

\paragraph{Primary Token Loads.}
\Cref{tab:primary-token-loads} reports mean Token Load for primary
configurations with complete token accounting, including the fixed system references.

% Generated by scripts/generate_resource_results.py; do not edit by hand.
\begin{table}[!htbp]
\centering
\small
\setlength{\tabcolsep}{4.5pt}
\renewcommand{\arraystretch}{1.08}
\caption{Mean primary Token Load (million tokens per scheduled attempt) for
configurations with complete token accounting. Dashes denote configurations
outside the evaluated panel. AgentSkillOS uses \texttt{skill\_200}. Fixed
references are reported as complete systems, separately from cross-harness indices.}
\label{tab:primary-token-loads}
\begin{tabular*}{\textwidth}{@{\extracolsep{\fill}}llcccccc@{}}
\toprule
Model & Harness & \makecell{SWE-\\bench-45} & \makecell{Terminal-\\Bench 2} & VitaBench & LOCA-Bench & ClawMark & \makecell{AgentSkill\\OS} \\
\midrule
\multirow{4}{*}{\makecell[l]{DeepSeek V4\\Flash Preview}} & DeepAgents & 2.358 & 2.484 & 0.933 & 6.181 & -- & -- \\
 & OpenCode & 1.531 & 0.677 & 0.503 & 2.390 & -- & -- \\
 & Pi CLI & 2.434 & 1.666 & 0.296 & 9.073 & -- & -- \\
 & Codex CLI & 3.762 & 1.244 & 0.656 & 1.288 & -- & -- \\
\midrule
\multirow{4}{*}{MiniMax M3} & DeepAgents & 6.444 & 1.860 & 0.562 & 6.332 & 6.333 & 2.950 \\
 & OpenCode & 10.722 & 1.882 & 0.404 & 14.644 & 3.563 & 3.679 \\
 & Pi CLI & 10.288 & 2.626 & 0.331 & 3.546 & 4.327 & 3.333 \\
 & Codex CLI & 3.850 & 2.826 & 0.724 & 3.105 & 7.709 & 38.040 \\
\midrule
\multirow{4}{*}{Kimi K2.7} & DeepAgents & 8.955 & 2.869 & 0.380 & 1.707 & 9.745 & 0.977 \\
 & OpenCode & 4.451 & 1.471 & 0.301 & 3.737 & 3.217 & 1.630 \\
 & Pi CLI & 4.849 & 0.834 & 0.229 & 3.280 & 0.056 & 1.567 \\
 & Codex CLI & 0.446 & 1.692 & 0.587 & 1.514 & 0.173 & 0.467 \\
\midrule
\multirow{4}{*}{\makecell[l]{Qwen3.6-\\35B-A3B}} & DeepAgents & 2.153 & 0.617 & 0.775 & 1.457 & 4.725 & 1.058 \\
 & OpenCode & 3.597 & 0.714 & 0.430 & 2.646 & 3.304 & 3.926 \\
 & Pi CLI & 19.018 & 1.316 & 0.277 & 3.537 & 3.197 & 3.274 \\
 & Codex CLI & 4.770 & 1.183 & 1.620 & 1.480 & 4.327 & 0.775 \\
\midrule
\multicolumn{2}{l}{GPT-5.5 xhigh + Codex CLI} & 2.197 & 0.927 & 0.461 & 2.193 & 6.510 & 1.285 \\
\cmidrule(lr){1-8}
\multicolumn{2}{l}{\makecell[l]{DeepSeek V4 Pro 0813\\+ DeepSeek Harness}} & 3.117 & 1.515 & 0.030 & 5.080 & -- & -- \\
\bottomrule
\end{tabular*}
\end{table}

\FloatBarrier

\paragraph{Complete configuration matrix.}
\Cref{tab:resource-cell-results} gives the complete resource matrix used
in \Cref{sec:results-token-efficiency}. It contains
\ResourceFullRows{} benchmark--condition--model--harness cells, including the
AgentSkillOS stress condition and all fixed references. Outcome and exact
token-coverage values are available for every cell.
For LOCA-Bench, task-local provider totals cover all 4,800 scheduled attempts;
tokens consumed before abnormal termination are retained.

\FloatBarrier
% Generated by scripts/generate_resource_results.py; do not edit by hand.
\begingroup
\scriptsize
\setlength{\tabcolsep}{3.2pt}
\renewcommand{\arraystretch}{0.90}
\begin{longtable}{@{}p{0.25\textwidth}p{0.16\textwidth}rrrr@{}}
\caption{Resource results by benchmark, condition, model, and harness.
Outcome: official Mean-of-4 or Task Score; Token Load: million tokens/attempt;
Tokens/Success: million tokens.
Coverage is the percentage of scheduled attempts with token accounting.}
\label{tab:resource-cell-results}\\
\toprule
Model & Harness & Outcome & \makecell{Token\\load} & \makecell{Tokens/\\success} & Coverage \\
\midrule
\endfirsthead
\multicolumn{6}{l}{\footnotesize\itshape Table~\thetable\ continued}\\
\toprule
Model & Harness & Outcome & \makecell{Token\\load} & \makecell{Tokens/\\success} & Coverage \\
\midrule
\endhead
\midrule
\multicolumn{6}{r}{\footnotesize Continued on next page}\\
\endfoot
\bottomrule
\endlastfoot
\addlinespace[4pt]
\multicolumn{6}{@{}l}{\textit{SWE-bench-45}}\\*[1pt]
DeepSeek V4 Flash Preview & DeepAgents & 31.11 & 2.358 & 7.579 & 100.00 \\
DeepSeek V4 Flash Preview & OpenCode & 32.22 & 1.531 & 4.750 & 100.00 \\
DeepSeek V4 Flash Preview & Pi CLI & 35.56 & 2.434 & 6.845 & 100.00 \\
DeepSeek V4 Flash Preview & Codex CLI & 36.11 & 3.762 & 10.419 & 100.00 \\
MiniMax M3 & DeepAgents & 31.11 & 6.444 & 20.714 & 100.00 \\
MiniMax M3 & OpenCode & 62.78 & 10.722 & 17.079 & 100.00 \\
MiniMax M3 & Pi CLI & 52.22 & 10.288 & 19.700 & 100.00 \\
MiniMax M3 & Codex CLI & 30.56 & 3.850 & 12.600 & 100.00 \\
Kimi K2.7 & DeepAgents & 15.56 & 8.955 & 57.566 & 100.00 \\
Kimi K2.7 & OpenCode & 48.89 & 4.451 & 9.103 & 100.00 \\
Kimi K2.7 & Pi CLI & 42.22 & 4.849 & 11.485 & 100.00 \\
Kimi K2.7 & Codex CLI & 10.56 & 0.446 & 4.225 & 100.00 \\
Qwen3.6-35B-A3B & DeepAgents & 15.00 & 2.153 & 14.352 & 100.00 \\
Qwen3.6-35B-A3B & OpenCode & 22.22 & 3.597 & 16.184 & 100.00 \\
Qwen3.6-35B-A3B & Pi CLI & 25.00 & 19.018 & 76.071 & 100.00 \\
Qwen3.6-35B-A3B & Codex CLI & 26.67 & 4.770 & 17.889 & 100.00 \\
\cmidrule(lr){1-6}
\multicolumn{2}{l}{GPT-5.5 xhigh + Codex CLI} & 65.56 & 2.197 & 3.351 & 100.00 \\
\cmidrule(lr){1-6}
\multicolumn{2}{l}{DeepSeek V4 Pro 0813 + DeepSeek Harness} & 70.56 & 3.117 & 4.418 & 100.00 \\
\addlinespace[4pt]
\multicolumn{6}{@{}l}{\textit{Terminal-Bench 2}}\\*[1pt]
DeepSeek V4 Flash Preview & DeepAgents & 45.79 & 2.484 & 5.425 & 100.00 \\
DeepSeek V4 Flash Preview & OpenCode & 40.45 & 0.677 & 1.674 & 100.00 \\
DeepSeek V4 Flash Preview & Pi CLI & 44.66 & 1.666 & 3.731 & 100.00 \\
DeepSeek V4 Flash Preview & Codex CLI & 37.64 & 1.244 & 3.306 & 100.00 \\
MiniMax M3 & DeepAgents & 21.63 & 1.860 & 8.598 & 100.00 \\
MiniMax M3 & OpenCode & 39.61 & 1.882 & 4.752 & 100.00 \\
MiniMax M3 & Pi CLI & 33.15 & 2.626 & 7.924 & 100.00 \\
MiniMax M3 & Codex CLI & 34.55 & 2.826 & 8.179 & 100.00 \\
Kimi K2.7 & DeepAgents & 30.62 & 2.869 & 9.369 & 100.00 \\
Kimi K2.7 & OpenCode & 39.04 & 1.471 & 3.767 & 100.00 \\
Kimi K2.7 & Pi CLI & 43.82 & 0.834 & 1.904 & 100.00 \\
Kimi K2.7 & Codex CLI & 31.74 & 1.692 & 5.331 & 100.00 \\
Qwen3.6-35B-A3B & DeepAgents & 26.40 & 0.617 & 2.337 & 100.00 \\
Qwen3.6-35B-A3B & OpenCode & 26.12 & 0.714 & 2.732 & 100.00 \\
Qwen3.6-35B-A3B & Pi CLI & 28.65 & 1.316 & 4.594 & 100.00 \\
Qwen3.6-35B-A3B & Codex CLI & 28.65 & 1.183 & 4.130 & 100.00 \\
\cmidrule(lr){1-6}
\multicolumn{2}{l}{GPT-5.5 xhigh + Codex CLI} & 60.39 & 0.927 & 1.535 & 100.00 \\
\cmidrule(lr){1-6}
\multicolumn{2}{l}{DeepSeek V4 Pro 0813 + DeepSeek Harness} & 46.91 & 1.515 & 3.229 & 100.00 \\
\addlinespace[4pt]
\multicolumn{6}{@{}l}{\textit{VitaBench}}\\*[1pt]
DeepSeek V4 Flash Preview & DeepAgents & 38.00 & 0.933 & 2.455 & 100.00 \\
DeepSeek V4 Flash Preview & OpenCode & 45.38 & 0.503 & 1.109 & 100.00 \\
DeepSeek V4 Flash Preview & Pi CLI & 48.94 & 0.296 & 0.605 & 100.00 \\
DeepSeek V4 Flash Preview & Codex CLI & 49.06 & 0.656 & 1.338 & 100.00 \\
MiniMax M3 & DeepAgents & 51.00 & 0.562 & 1.102 & 100.00 \\
MiniMax M3 & OpenCode & 49.38 & 0.404 & 0.819 & 100.00 \\
MiniMax M3 & Pi CLI & 49.25 & 0.331 & 0.672 & 100.00 \\
MiniMax M3 & Codex CLI & 45.06 & 0.724 & 1.607 & 100.00 \\
Kimi K2.7 & DeepAgents & 43.06 & 0.380 & 0.883 & 100.00 \\
Kimi K2.7 & OpenCode & 44.38 & 0.301 & 0.678 & 100.00 \\
Kimi K2.7 & Pi CLI & 40.25 & 0.229 & 0.569 & 100.00 \\
Kimi K2.7 & Codex CLI & 41.44 & 0.587 & 1.416 & 100.00 \\
Qwen3.6-35B-A3B & DeepAgents & 27.69 & 0.775 & 2.799 & 100.00 \\
Qwen3.6-35B-A3B & OpenCode & 30.88 & 0.430 & 1.393 & 100.00 \\
Qwen3.6-35B-A3B & Pi CLI & 31.19 & 0.277 & 0.887 & 100.00 \\
Qwen3.6-35B-A3B & Codex CLI & 22.88 & 1.620 & 7.081 & 100.00 \\
\cmidrule(lr){1-6}
\multicolumn{2}{l}{GPT-5.5 xhigh + Codex CLI} & 46.56 & 0.461 & 0.989 & 100.00 \\
DeepSeek V4 Pro 0813 & \makecell[l]{DeepSeek\\Harness} & 54.13 & 0.030 & 0.056 & 100.00 \\
\addlinespace[4pt]
\multicolumn{6}{@{}l}{\textit{LOCA-Bench}}\\*[1pt]
DeepSeek V4 Flash Preview & DeepAgents & 20.67 & 6.181 & 29.910 & 100.00 \\
DeepSeek V4 Flash Preview & OpenCode & 31.00 & 2.390 & 7.709 & 100.00 \\
DeepSeek V4 Flash Preview & Pi CLI & 29.00 & 9.073 & 31.287 & 100.00 \\
DeepSeek V4 Flash Preview & Codex CLI & 50.67 & 1.288 & 2.542 & 100.00 \\
MiniMax M3 & DeepAgents & 24.00 & 6.332 & 26.384 & 100.00 \\
MiniMax M3 & OpenCode & 21.00 & 14.644 & 69.736 & 100.00 \\
MiniMax M3 & Pi CLI & 12.67 & 3.546 & 27.992 & 100.00 \\
MiniMax M3 & Codex CLI & 87.00 & 3.105 & 3.569 & 100.00 \\
Kimi K2.7 & DeepAgents & 8.33 & 1.707 & 20.482 & 100.00 \\
Kimi K2.7 & OpenCode & 27.67 & 3.737 & 13.506 & 100.00 \\
Kimi K2.7 & Pi CLI & 34.33 & 3.280 & 9.553 & 100.00 \\
Kimi K2.7 & Codex CLI & 54.00 & 1.514 & 2.803 & 100.00 \\
Qwen3.6-35B-A3B & DeepAgents & 8.67 & 1.457 & 16.810 & 100.00 \\
Qwen3.6-35B-A3B & OpenCode & 21.00 & 2.646 & 12.601 & 100.00 \\
Qwen3.6-35B-A3B & Pi CLI & 15.00 & 3.537 & 23.578 & 100.00 \\
Qwen3.6-35B-A3B & Codex CLI & 36.33 & 1.480 & 4.073 & 100.00 \\
\cmidrule(lr){1-6}
\multicolumn{2}{l}{GPT-5.5 xhigh + Codex CLI} & 76.00 & 2.193 & 2.885 & 100.00 \\
\cmidrule(lr){1-6}
\multicolumn{2}{l}{DeepSeek V4 Pro 0813 + DeepSeek Harness} & 73.33 & 5.080 & 6.927 & 100.00 \\
\addlinespace[4pt]
\multicolumn{6}{@{}l}{\textit{ClawMark}}\\*[1pt]
MiniMax M3 & DeepAgents & 54.78 & 6.333 & -- & 100.00 \\
MiniMax M3 & OpenCode & 69.62 & 3.563 & -- & 100.00 \\
MiniMax M3 & Pi CLI & 62.58 & 4.327 & -- & 100.00 \\
MiniMax M3 & Codex CLI & 58.49 & 7.709 & -- & 100.00 \\
Kimi K2.7 & DeepAgents & 58.65 & 9.745 & -- & 100.00 \\
Kimi K2.7 & OpenCode & 70.36 & 3.217 & -- & 100.00 \\
Kimi K2.7 & Pi CLI & 11.93 & 0.056 & -- & 100.00 \\
Kimi K2.7 & Codex CLI & 11.96 & 0.173 & -- & 100.00 \\
Qwen3.6-35B-A3B & DeepAgents & 52.71 & 4.725 & -- & 100.00 \\
Qwen3.6-35B-A3B & OpenCode & 63.62 & 3.304 & -- & 100.00 \\
Qwen3.6-35B-A3B & Pi CLI & 52.61 & 3.197 & -- & 100.00 \\
Qwen3.6-35B-A3B & Codex CLI & 61.21 & 4.327 & -- & 100.00 \\
\cmidrule(lr){1-6}
\multicolumn{2}{l}{GPT-5.5 xhigh + Codex CLI} & 74.31 & 6.510 & -- & 100.00 \\
\addlinespace[4pt]
\multicolumn{6}{@{}l}{\textit{AgentSkillOS (\texttt{skill\_200}, primary)}}\\*[1pt]
MiniMax M3 & DeepAgents & 80.07 & 2.950 & -- & 100.00 \\
MiniMax M3 & OpenCode & 93.49 & 3.679 & -- & 100.00 \\
MiniMax M3 & Pi CLI & 85.39 & 3.333 & -- & 100.00 \\
MiniMax M3 & Codex CLI & 36.44 & 38.040 & -- & 100.00 \\
Kimi K2.7 & DeepAgents & 89.92 & 0.977 & -- & 100.00 \\
Kimi K2.7 & OpenCode & 84.94 & 1.630 & -- & 100.00 \\
Kimi K2.7 & Pi CLI & 95.91 & 1.567 & -- & 100.00 \\
Kimi K2.7 & Codex CLI & 55.66 & 0.467 & -- & 100.00 \\
Qwen3.6-35B-A3B & DeepAgents & 74.01 & 1.058 & -- & 100.00 \\
Qwen3.6-35B-A3B & OpenCode & 76.12 & 3.926 & -- & 100.00 \\
Qwen3.6-35B-A3B & Pi CLI & 56.23 & 3.274 & -- & 100.00 \\
Qwen3.6-35B-A3B & Codex CLI & 63.06 & 0.775 & -- & 100.00 \\
\cmidrule(lr){1-6}
\multicolumn{2}{l}{GPT-5.5 xhigh + Codex CLI} & 97.57 & 1.285 & -- & 100.00 \\
\addlinespace[4pt]
\multicolumn{6}{@{}l}{\textit{AgentSkillOS (\texttt{skill\_900}, stress)}}\\*[1pt]
MiniMax M3 & DeepAgents & 83.02 & 4.054 & -- & 100.00 \\
MiniMax M3 & OpenCode & 64.08 & 6.475 & -- & 100.00 \\
MiniMax M3 & Pi CLI & 89.98 & 4.079 & -- & 100.00 \\
MiniMax M3 & Codex CLI & 36.42 & 9.668 & -- & 100.00 \\
Kimi K2.7 & DeepAgents & 84.34 & 2.128 & -- & 100.00 \\
Kimi K2.7 & OpenCode & 62.87 & 2.576 & -- & 100.00 \\
Kimi K2.7 & Pi CLI & 90.87 & 3.031 & -- & 100.00 \\
Kimi K2.7 & Codex CLI & 66.57 & 0.425 & -- & 100.00 \\
Qwen3.6-35B-A3B & DeepAgents & 61.77 & 3.044 & -- & 100.00 \\
Qwen3.6-35B-A3B & OpenCode & 50.15 & 4.123 & -- & 100.00 \\
Qwen3.6-35B-A3B & Pi CLI & 75.98 & 4.536 & -- & 100.00 \\
Qwen3.6-35B-A3B & Codex CLI & 71.42 & 0.574 & -- & 100.00 \\
\cmidrule(lr){1-6}
\multicolumn{2}{l}{GPT-5.5 xhigh + Codex CLI} & 89.68 & 0.906 & -- & 100.00 \\
\end{longtable}
\endgroup

\end{document}